\documentclass[11pt]{article}

\usepackage[final]{acl}

\usepackage{times}
\usepackage{latexsym}
\usepackage[T1]{fontenc}
\usepackage[utf8]{inputenc}
\usepackage{microtype}
\usepackage{inconsolata}
\usepackage{graphicx}
\usepackage{amsmath,amssymb,bbm}
\usepackage{algorithm}
\usepackage{algorithmic}
\usepackage{booktabs}
\usepackage{multirow}
\usepackage{xcolor}
\usepackage{enumitem}
\usepackage[normalem]{ulem}
\usepackage{pgfplots}
\pgfplotsset{compat=1.18}
\usepgfplotslibrary{groupplots}
\definecolor{mctsdark}{RGB}{90,90,90}
\definecolor{mctslight}{RGB}{160,160,160}
\definecolor{abmctsblue}{RGB}{55,126,184}
\definecolor{oursred}{RGB}{228,26,28}

\definecolor{placeholder}{gray}{0.50}

\colorlet{revcolor}{black}
\newcommand{\rev}[1]{{\color{revcolor}#1}}
\newcommand{\ours}{ExTS}

\title{Exploit More, Explore Smarter for Budget-Constrained Agentic Search}

\author{Haoyang Fang \\
  Amazon AGI \\
  \texttt{haoyfang@amazon.com} \\ \And
  Bernie Wang\footnotemark[1] \\
  Amazon AGI \\
  \texttt{yuyawang@amazon.com} \\}

\begin{document}
\renewcommand{\thefootnote}{\fnsymbol{footnote}}
\maketitle
\footnotetext[1]{Work done at Amazon.}
\renewcommand{\thefootnote}{\arabic{footnote}}
\setcounter{footnote}{0}

% ============================================================================
% ABSTRACT
% ============================================================================
\begin{abstract}
Budget-constrained agentic search arises when an LLM agent must refine candidates under a small evaluation budget, because validation is expensive, generation requires multiple model calls, or both. In this regime, standard MCTS allocates budget poorly: exploration bonuses dominate at low visit counts, unpromising siblings are expanded before promising chains can deepen, and branching is independent of node quality. We introduce \textbf{\ours{}}, a tree-search policy that treats expansion itself as a value-of-information decision. \ours{} combines three mechanisms: discriminative reward shaping to separate candidates under narrow score distributions, a stochastic virtual child that estimates the value of creating a new branch from the parent's reward history, and quality-conditioned branching that expands only when a node's score justifies the budget cost. Across prompt optimization, code generation, molecular structure elucidation, and agentic workflow optimization, \ours{} is competitive with or improves over task-specific tree-search baselines, with an average relative gain of \textbf{+5.5\%} using a single fixed configuration. We further introduce pilot-run diagnostics that characterize what makes budget-constrained agentic search problems structurally different from one another, providing both understanding of the problem space and practical guidance for adaptation.
\end{abstract}

\begin{figure*}[t]
\centering
\includegraphics[width=0.85\textwidth]{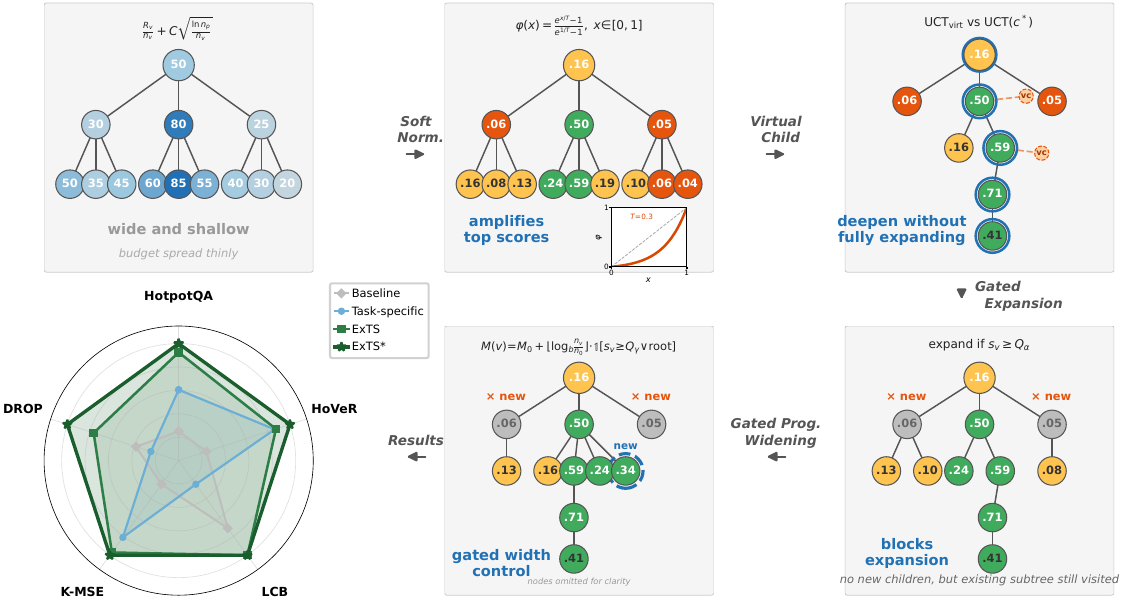}
\caption{Standard UCT spreads budget across a wide, shallow tree via flat rewards and obligatory expansion. \ours{} instead uses virtual children that frame expansion as a value-of-information decision and quality-conditioned branching gated on the reshaped score. Radar plot: normalized gains; \ours{}$^*$ denotes pilot-adapted configurations.}
\label{fig:teaser}
\end{figure*}

% ============================================================================
% 1. INTRODUCTION
% ============================================================================
\section{Introduction}

A growing class of AI systems rely on \emph{agentic search}: an LLM iteratively proposes and refines candidates, consuming budget on both generation and validation. These systems span prompt optimization \citep{agrawal2025gepa, opsahlong2024miprov2}, code generation \citep{zhang2023planning, inoue2025abmcts}, tool-augmented reasoning \citep{yao2023react, schick2023toolformer}, multi-step planning \citep{yao2023tot, besta2024got}, etc. Despite their diversity, they share a common constraint: the search budget is limited to tens or hundreds of calls. We call this regime \emph{budget-constrained agentic search}.

Recent work on LLM search focuses predominantly on the \emph{harness}: the framework wrapping the search, such as value functions, action spaces, or multi-agent evaluation \citep{shinn2023reflexion, madaan2023selfrefine, zhou2024lats, qi2024rstar, antoniades2024swesearch, zhang2024llamaberry, hao2023rap, mlmaster, sstar, mlzero}. These systems retain linear search or standard UCT for selection and expansion (Appendix~\ref{sec:related}), leaving substantial room for improvement in how budget is allocated within the tree. On the search algorithm side, AB-MCTS \citep{inoue2025abmcts} replaces UCT with Thompson sampling, GEPA \citep{agrawal2025gepa} uses Pareto-frontier selection, and AFlow \citep{aflow} applies score-weighted random sampling, but each targets a single task type and none condition expansion on node quality or shape rewards for narrow score distributions. How to design selection and expansion policies that transfer across diverse agentic problems has received limited attention.

We introduce \textbf{\ours{}}\footnote{\rev{Code will be released at \url{https://github.com/amazon-science/ExTS}. Until then, please contact us.}}, a tree-search policy that \emph{exploits more and explores smarter} by jointly redesigning both \emph{how nodes are selected} and \emph{whether expansion should occur}. The core insight is that under tight budgets, the decision of whether to expand is as consequential as which child to visit. Standard MCTS suffers from three structural limitations in this regime: exploration-bonus dominance at low visit counts, forced full expansion of unpromising siblings, and no quality gate on branching. \ours{} addresses each through three mechanisms: \textbf{discriminative reward shaping} (Section~\ref{sec:shaping}) that produces effective selection signals when raw scores cluster narrowly; a \textbf{virtual child heuristic} (Section~\ref{sec:virtual}) that estimates expansion value by sampling from the parent's reward history, making expansion a value-of-information (VOI) decision competing directly with deepening; and \textbf{quality-conditioned branching} (Section~\ref{sec:scoregate}) that restricts expansion to nodes whose score justifies the budget cost. These mechanisms are inspired by classical MCTS ideas (progressive widening \citep{coulom2007computing}, first-play urgency \citep{gelly2006fpu}, PUCT \citep{rosin2011puct}) and improve upon each for budget-constrained agentic search (Section~\ref{sec:ablations}).

With a single fixed configuration across prompt optimization \citep{yang2018hotpotqa, jiang2020hover}, code generation \citep{jain2024livecodebench}, molecular structure elucidation \citep{kmse}, and agentic workflow optimization \citep{dua2019drop}, \ours{} is competitive with or improves over task-specific methods (+10.8\% on HotpotQA, +11.7\% on LiveCodeBench hard, +1.3\% on K-MSE, +3.5\% on DROP, and +0.2\% on HoVeR). \ours{}$^*$ further improves performance by adjusting one or \rev{two} hyperparameters based on pilot-run diagnostics. We additionally report diagnostics linking landscape structure to hyperparameter sensitivity, component ablations, budget-scaling and tree-shape analysis.

% ============================================================================
% 2. PRELIMINARIES
% ============================================================================
\section{Preliminaries}

\subsection{Validation-Heavy Search}
\label{sec:validation-heavy}

We formalize validation-heavy search as a tuple $(\mathcal{S}, \mathcal{A}, f, B)$ where $\mathcal{S}$ is the set of candidate solutions (e.g., prompts, code, or plans), $\mathcal{A}: \mathcal{S} \rightarrow \mathcal{S}$ is a stochastic refinement operator powered by an LLM, $f: \mathcal{S} \rightarrow \mathbb{R} \cup \{\bot\}$ is a validation function that returns a score or failure, and $B$ is the total evaluation budget. The search builds a tree $\mathcal{T}$ rooted at an initial candidate $x_0$ and the goal is $\arg\max_{x \in \mathcal{T}} f(x)$ within budget $B$.

We identify four \emph{pilot-run diagnostics} that characterize the search landscape of a validation-heavy task for a given model and scorer (formal definitions in Appendix~\ref{app:properties}). Beyond guiding hyperparameter choices, these diagnostics characterize the axes along which agentic search problems structurally differ and help explain why no single search configuration dominates universally. These diagnostics are computed from a pilot tree built by running the \emph{baseline} search method for each dataset, measuring properties of the landscape as seen by the task-native algorithm. The \emph{refinement variance} $\hat{\sigma}_{\mathcal{A}}$ is the normalized standard deviation of $f$ across independent refinements with the same input, quantifying the stochasticity of the LLM refinement operator. The \emph{normalized score deviation} $\hat{\sigma}_f$ is the standard deviation of $f$ across all successfully validated candidates in the completed tree, normalized to $[0,1]$. The \emph{score drift} $\kappa$ is the mean absolute shift in normalized scores caused by evolving normalization bounds as new candidates are discovered; high drift indicates that the search frequently discovers candidates at the extremes of the score range. The \emph{failure rate} $\rho$ is the fraction of refinement attempts where $f$ returns $\bot$. Table~\ref{tab:domains} characterizes the datasets we study along these axes.

\begin{table}[t]
\caption{Pilot-run diagnostics (mean$\pm$std, 3 seeds). $\hat{\sigma}_{\mathcal{A}}$: refinement variance; $\hat{\sigma}_f$: score spread; $\kappa$: score drift; $\rho$: failure rate.}
\label{tab:domains}
\centering
\small
\resizebox{\columnwidth}{!}{%
\begin{tabular}{lcccc}
\toprule
\textbf{Dataset} & $\hat{\sigma}_{\mathcal{A}}$ & $\hat{\sigma}_f$ & $\kappa$ & $\rho$ \\
\midrule
HotpotQA & .30\tiny{$\pm$.05} & .30\tiny{$\pm$.09} & .11\tiny{$\pm$.03} & .74\tiny{$\pm$.04} \\
HoVeR & .20\tiny{$\pm$.06} & .29\tiny{$\pm$.01} & .16\tiny{$\pm$.11} & .82\tiny{$\pm$.01} \\
LiveCodeBench & .32\tiny{$\pm$.01} & .23\tiny{$\pm$.01} & .14\tiny{$\pm$.13} & .57\tiny{$\pm$.05} \\
K-MSE & .09\tiny{$\pm$.00} & .18\tiny{$\pm$.00} & .05\tiny{$\pm$.00} & $<$.01 \\
DROP & .04\tiny{$\pm$.03} & .23\tiny{$\pm$.03} & .02\tiny{$\pm$.01} & .03\tiny{$\pm$.03} \\
\bottomrule
\end{tabular}%
}
\end{table}

\subsection{Standard MCTS and Its Limitations}
\label{sec:uct-limits}

In standard MCTS, the UCT policy \citep{kocsis2006uct} selects child $c$ of node $v$ by maximizing
\begin{equation}
\text{UCT}_{\text{std}}(c) = \frac{R_c}{n_c} + C \sqrt{\frac{\ln n_v}{n_c}}
\end{equation}
where $R_c/n_c$ is the average reward, $n_c$ and $n_v$ are visit counts, and $C$ controls exploration. The theoretical default $C = \sqrt{2}$ is calibrated for rewards in $[0, 1]$ with sufficient budget for convergence \citep{browne2012mcts}.

Two assumptions underlying UCT are violated in validation-heavy search. First, reward shaping (Section~\ref{sec:shaping}) compresses the effective exploitation range, causing the exploration bonus to dominate and making all nodes appear equally promising. Second, budgets of tens to hundreds of evaluations are far below the asymptotic regime where UCT's convergence guarantees hold \citep{silver2016alphago, browne2012mcts, kocsis2006uct}. Together, these violations produce flat, wide trees that fail to develop deep refinement chains. The PUCT exploration term (Section~\ref{sec:shaping}) partially addresses the first issue by decaying linearly rather than logarithmically, reducing the dominance of exploration under tight budgets.

\paragraph{Classical extensions.} Three classical MCTS ideas are relevant but do not transfer directly to this regime (Section~\ref{sec:ablations}). Progressive widening \citep{coulom2007computing, chaslot2008progressive} ties branching to visit count but is score-agnostic. First-play urgency (FPU) \citep{gelly2006fpu} assigns a fixed value to unvisited actions but cannot adapt to non-stationary score distributions. PUCT \citep{rosin2011puct, silver2017alphazero} uses a learned policy prior with linear exploration decay. \ours{} redesigns each: progressive widening becomes quality-conditioned, FPU becomes a stochastic virtual child sampling from the parent's reward history, and PUCT-style decay operates without a learned prior.

% ============================================================================
% 3. METHODOLOGY
% ============================================================================
\section{Methodology}

{\color{revcolor} \emph{Our intuition is that, especially under tight budgets, adding a single scalar-weighted UCT or PUCT exploration term to the reward cannot by itself capture the balance between exploration and exploitation (Section~\ref{sec:exploration-constant}). \ours{} keeps this weighted exploration term, but conditions both selection and expansion decisions on the observed reward distribution at two scopes: local (a node's own subtree) and global (the whole search tree).}}

Concretely, each node $v$ stores a candidate $x_v \in \mathcal{S}$ and maintains: $n_v$ (total visits), $n_v^+$ (successful visits), $\mathcal{R}_v$ (reward observations from successful expansions in $v$'s subtree), and $s_v = f(x_v)$ (validation score). The reward pool $\mathcal{R}_v$ implements mean backup: ExUCT$(v)$ estimates the refinement productivity of expanding below $v$, not the quality of $v$'s own candidate. Figure~\ref{fig:teaser} illustrates how the three design choices transform the search tree. The complete search loop follows standard MCTS (select, expand, validate, backpropagate) and is given in Algorithm~\ref{alg:exts} (Appendix~\ref{app:defaults}).

\subsection{Discriminative Reward Shaping}
\label{sec:shaping}

Both effective selection and meaningful expansion decisions require differentiating node quality. When raw validation scores cluster in a narrow range, as is common in validation-heavy domains, UCT's exploitation term makes all nodes appear equally attractive: selection becomes near-random, and any VOI comparison between expanding versus deepening becomes uninformative. The \ours{} selection policy addresses this by scoring each visited node $v$ ($n_v > 0$) as:
\begin{align}
\label{eq:uct}
\text{ExUCT}(v) &= \left(\frac{n_v^+}{n_v}\right)^\alpha \!\cdot\! \overline{\varphi(r_i)} \nonumber\\
&\quad + \text{explore}(n_{\mathrm{par}(v)},\; n_v)
\end{align}
The exploitation term combines two signals: the shaped rewards $\overline{\varphi(r_i)}$, which separate candidates when raw scores cluster narrowly, and a success-rate weight $(n_v^+/n_v)^\alpha$, which discounts nodes whose subtrees produce mostly invalid outputs ($\alpha$ controls failure discounting). The shaping function $\varphi$ maps raw scores through \emph{global} normalization and a temperature-controlled nonlinearity:
\begin{equation}
\label{eq:shaping}
\varphi(r) = \frac{e^{\,\hat{r}/T} - 1}{e^{1/T} - 1}, \quad \hat{r} = \frac{r - s_{\min}}{s_{\max} - s_{\min}},
\end{equation}
\begin{equation}
\overline{\varphi(r_i)} = \frac{1}{|\mathcal{R}_v|}\sum_{r_i \in \mathcal{R}_v} \varphi(r_i).
\end{equation}
For $T < 1$, $\varphi$ is convex, compressing low scores and amplifying high scores. This is important in validation-heavy domains where scores cluster in a narrow range and linear normalization would make most nodes appear equally attractive.

The exploration term takes one of two forms. The \textbf{UCT} variant decays logarithmically:
\begin{equation}
\label{eq:uct-explore}
\text{explore}_{\text{UCT}}(n_{\text{par}}, n_v) = C \sqrt{\frac{\ln n_{\text{par}}}{n_v}},
\end{equation}
while the \textbf{PUCT-style} variant decays linearly, providing faster convergence toward exploitation under tight budgets:
\begin{equation}
\label{eq:puct-explore}
\text{explore}_{\text{PUCT}}(n_{\text{par}}, n_v) = C \cdot \frac{\sqrt{n_{\text{par}}}}{1 + n_v}.
\end{equation}
Unlike standard PUCT \citep{rosin2011puct, silver2017alphazero}, which incorporates a learned policy prior $P(a|s)$, our variant uses no prior for simplicity, because such a prior must be designed task-specifically. The default \ours{} configuration uses PUCT-style exploration with $C\!=\!1.0$; we evaluate the UCT variant ($C\!=\!\sqrt{2}$) as an ablation in Section~\ref{sec:ablations}.

\subsection{VOI Estimation via the Virtual Child}
\label{sec:virtual}

The core of \ours{}'s expansion-as-VOI principle is a mechanism for estimating the expected value of creating a new child at any internal node. In standard MCTS, expansion occurs only at leaf nodes, forcing the algorithm to fully expand each level before deepening. We instead make expansion a first-class decision at every internal node by introducing a \emph{virtual child} that competes with real children during selection.

At each internal node $v$ with $m$ existing children, if $v$ can still expand ($m < M$), we compute a virtual child score. First, we estimate the virtual child's fair-share visit count:
\begin{equation}
\label{eq:nfair}
n_{\text{fair}} = \max\!\left(1,\; {\color{revcolor}\frac{\textstyle\sum_{c \in \mathrm{children}(v)} n_c}{m}}\right).
\end{equation}
We then sample $k = \max(1, \lfloor n_{\text{fair}} \cdot n_v^+/n_v \rfloor)$ scores with replacement from the parent's reward pool $\mathcal{R}_v \cup \{s_v\}$ and compute the estimated shaped mean $\overline{\varphi(\tilde{r})}$. The virtual child score mirrors Equation~\ref{eq:uct}:
\begin{equation}
\label{eq:virtual}
\text{ExUCT}_{\text{virt}} = \Bigl(\frac{n_v^+}{n_v}\Bigr)^{\!\alpha} \!\cdot\, \overline{\varphi(\tilde{r})} \;+\; \text{explore}(n_v, n_{\text{fair}}).
\end{equation}
The stochastic sampling captures finite-sample uncertainty: fewer expected visits produce fewer samples and thus higher variance, naturally encouraging expansion when evidence is thin.

Let $c^* = \arg\max_{c_j} \text{ExUCT}(c_j)$ be the best real child. If $\text{ExUCT}_{\text{virtual}} > \text{ExUCT}(c^*)$, the node is selected for expansion; otherwise selection recurses into $c^*$. \rev{Appendix~\ref{app:virtual-example} works through this comparison on a real run.}

\subsection{Quality-Conditioned Branching}
\label{sec:scoregate}
\label{sec:widening}

\ours{} conditions branching on node quality at two granularities: whether a node may expand at all, and how many children it may accumulate.

\paragraph{Expansion gate.} For non-root nodes, expansion is permitted only when the node's validation score exceeds the $\tau$-quantile of the population:
\begin{equation}
s_v \geq Q_\tau\bigl(\{s_u : u \in \mathcal{T}\}\bigr).
\end{equation}
Below-threshold nodes remain reachable and update their statistics but cannot spawn new children. The threshold adapts as the tree accumulates better candidates, becoming increasingly selective over time.

\paragraph{Quality-gated progressive widening.} Classical progressive widening \citep{coulom2007computing, chaslot2008progressive} ties branching to visit count but is score-agnostic. We condition it on node quality: the maximum branching factor grows with visit count, but only for nodes whose quality justifies it:
\begin{equation}
\label{eq:widening}
M(v) = M_0 + \bigl\lfloor \log_b\!\tfrac{n_v}{n_0} \bigr\rfloor \cdot \mathbbm{1}\bigl[s_v \geq Q_\gamma \lor v = v_{\text{root}}\bigr],
\end{equation}
where $M_0$ is the initial maximum children, $b$ is the widening base, $n_0$ is the visit threshold before widening begins, and $Q_\gamma$ is the $\gamma$-quantile of node scores. Typically $\gamma \gg \tau$ (defaults: $\gamma = 0.75$, $\tau = 0.25$), reflecting that widening is a stronger commitment than a single expansion. Low-scoring nodes remain capped at $M_0$.

\begin{algorithm}[t]
\caption{\textsc{ExUCT-Select}: Selection with Virtual Child}
\label{alg:uct-select}
\label{sec:defaults}
\begin{algorithmic}[1]
\REQUIRE Node $v$
\IF{$v$ is a leaf}
    \RETURN $v$
\ENDIF
\STATE $\text{can\_expand} \leftarrow |\text{children}(v)| < M(v)$
\IF{$v \neq \text{root}$ \AND can\_expand}
    \STATE can\_expand $\leftarrow s_v \geq Q_\tau(\{s_u : u \in \mathcal{T}\})$
\ENDIF
\STATE $c^* \leftarrow \arg\max_{c \in \text{children}(v)} \text{ExUCT}(c)$
\IF{can\_expand}
    \STATE $m \leftarrow |\text{children}(v)|$
    \STATE $n_{\text{fair}} \leftarrow \max\bigl(1,\; {\color{revcolor}\sum_{c} n_c / m}\bigr)$
    \STATE $k \leftarrow \max(1, \lfloor n_{\text{fair}} \cdot n_v^+/n_v \rfloor)$
    \STATE Sample $\tilde{r}_1, \ldots, \tilde{r}_k \sim \mathcal{R}_v \cup \{s_v\}$ with replacement
    \STATE $q \leftarrow (n_v^+/n_v)^\alpha \cdot \frac{1}{k}\sum_i \varphi(\tilde{r}_i)$
    \STATE $\text{ExUCT}_{\text{virtual}} \leftarrow q + \text{explore}(n_v, n_{\text{fair}})$
    \IF{$\text{ExUCT}_{\text{virtual}} > \text{ExUCT}(c^*)$}
        \RETURN $v$ \hfill {\small expand here}
    \ENDIF
\ENDIF
\RETURN \textsc{ExUCT-Select}($c^*$)
\end{algorithmic}
\end{algorithm}

\paragraph{Gate semantics.} The expansion gate (Algorithm~\ref{alg:uct-select}, line~5) applies to \emph{internal} nodes deciding whether to create additional children. A leaf node returned at line~2 is selected for its first expansion regardless of score; the gate restricts further branching only after a node has at least one child and evidence of its quality. This means every node receives at least one expansion attempt before gating takes effect.

% ============================================================================
% 4. EXPERIMENTS
% ============================================================================
\section{Evaluation}
\label{sec:setup}

We evaluate \ours{} across four domains spanning diverse failure rates, score variances, and drift levels (Table~\ref{tab:domains}). All methods share the same evaluation budget $B$ per task. \ours{} uses a single fixed configuration across all domains (Table~\ref{tab:defaults}); each baseline is the best-known method for its respective task with a fully tuned, domain-specific configuration. We additionally report \ours{}$^*$ variants that adjust one or two hyperparameters per task based on pilot-run diagnostics, a modest adaptation that remains fair given the domain-specific tuning already present in each baseline. Full formulation details are in Appendix~\ref{app:setup}.

\subsection{Prompt Optimization}
\label{sec:exp-prompt}

\begin{table}[t]
\caption{Prompt optimization: test accuracy (\%) over 3 seeds. $^\dagger$Results from \citet{agrawal2025gepa}. $^*$Pilot-adapted (Section~\ref{sec:exploit-explore}).}
\label{tab:prompt-main}
\centering
\small
\begin{tabular}{lcc}
\toprule
\textbf{Method} & \textbf{HotpotQA} & \textbf{HoVeR} \\
\midrule
Baseline & 41.44\tiny{$\pm$1.25} & 36.11\tiny{$\pm$0.79} \\
GRPO$^\dagger$ & 43.33 & 38.67 \\
MIPROv2$^\dagger$ & 55.33 & 47.33 \\
GEPA (Pareto) & 58.55\tiny{$\pm$5.32} & \underline{50.33}\tiny{$\pm$2.45} \\
\ours{} & \textbf{64.89}\tiny{$\pm$0.83} & \underline{50.45}\tiny{$\pm$2.18} \\
\ours{}$^*$ & \textbf{66.00}\tiny{$\pm$0.98} & \textbf{51.67}\tiny{$\pm$0.98} \\
\bottomrule
\end{tabular}
\end{table}

We integrate \ours{} into the GEPA framework \citep{agrawal2025gepa} for optimizing DSPy \citep{khattab2024dspy} instructions on HotpotQA \citep{yang2018hotpotqa} and HoVeR \citep{jiang2020hover} (300 test examples, 3 seeds, Qwen3-8B). This domain has the highest failure rate ($\rho = 0.74$--$0.82$) among our tasks.

Table~\ref{tab:prompt-main} shows that \ours{} improves over GEPA Pareto by $+$10.8\% on HotpotQA with $6.4\times$ lower variance. The pilot-adapted \ours{}$^*$ further improves to 66.00\% by increasing the success-rate exponent to $\alpha\!=\!2.0$, which sharpens node selection in this high-failure-rate domain ($\rho = 0.74$; Section~\ref{sec:exploit-explore}). On HoVeR, \ours{} performs on par with GEPA Pareto ($+$0.2\%, within noise). HoVeR exhibits the highest score drift ($\kappa = 0.16$) among our tasks, causing normalization bounds to shift frequently and re-rank nodes mid-search. The pilot-adapted \ours{}$^*$ switches to UCT exploration with no gating, buffering against score re-ranking by maintaining sustained exploration (Section~\ref{sec:exploit-explore}).

\subsection{Code Generation}
\label{sec:exp-code}

\begin{figure}[t]
\centering
\includegraphics[width=\columnwidth]{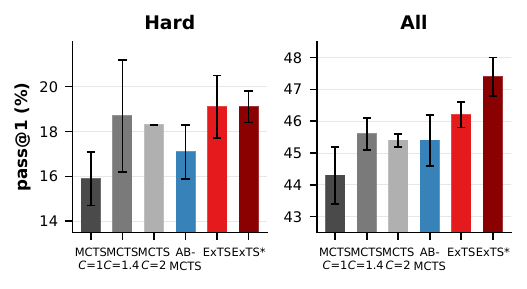}
\caption{LiveCodeBench pass@1 (\%) over 3 seeds.}
\label{fig:code-main}
\end{figure}

We integrate \ours{} into the TreeQuest framework \citep{inoue2025abmcts} on LiveCodeBench \citep{jain2024livecodebench} (182 problems, budget 128, Claude Sonnet 4), comparing against StandardMCTS and AB-MCTS-A. All three methods achieve identical public-test success rates on easy (100\%) and medium (85.8\%), making those splits non-discriminative; we therefore use the hard split as our primary metric while reporting overall performance for reference (Figure~\ref{fig:code-main}).

On the hard split, \ours{} reaches 19.1\% pass@1 versus 17.1\% for AB-MCTS-A ($+$11.7\%) and 18.1\% for StandardMCTS ($+$5.5\%), with the lowest cross-seed variance. The gains come from deeper refinement of promising candidates rather than broader coverage. Overall pass@1 is 46.2\% ($+$1.3\% over StandardMCTS). The pilot-adapted \ours{}$^*$ ($T\!=\!0.5$) further improves to 47.4\% overall ($+$4.0\% over StandardMCTS).

\subsection{Molecular Structure Elucidation}
\label{sec:exp-mol}

\begin{table}[t]
\caption{K-MSE molecular elucidation (Claude Sonnet 4.6, 216 molecules, 16 rollouts, 3 seeds). ACC (exact match) is the primary metric; fingerprint similarities are listed for reference. $^*$Pilot-adapted (Section~\ref{sec:exploit-explore}).}
\label{tab:mol-main}
\centering
\small
\resizebox{\columnwidth}{!}{%
\begin{tabular}{lcccc}
\toprule
\textbf{Method} & \textbf{ACC (\%)} & \textbf{Morgan} & \textbf{MACCS} & \textbf{RDK} \\
\midrule
CoT & 79.8\tiny{$\pm$0.2} & .887\tiny{$\pm$.002} & .942\tiny{$\pm$.001} & .904\tiny{$\pm$.001} \\
K-MSE & 89.8\tiny{$\pm$0.4} & .952\tiny{$\pm$.002} & .976\tiny{$\pm$.001} & .962\tiny{$\pm$.003} \\
\ours{} & \textbf{91.0}\tiny{$\pm$0.6} & .952\tiny{$\pm$.004} & .972\tiny{$\pm$.005} & .960\tiny{$\pm$.004} \\
\ours{}$^*$ & \textbf{91.2}\tiny{$\pm$0.4} & .954\tiny{$\pm$.004} & .973\tiny{$\pm$.003} & .959\tiny{$\pm$.004} \\
\bottomrule
\end{tabular}%
}
\end{table}

We apply \ours{} to K-MSE \citep{kmse}, where an LLM deduces molecular SMILES from NMR and IR spectra (216 molecules, Claude Sonnet 4.6). This domain has near-zero failure rate and stationary scores.

\ours{} improves ACC by $+$1.3\% over K-MSE's MCTSr (Table~\ref{tab:mol-main}). The strong base model leaves limited headroom; gains come from exploitation shaping and the virtual child directing budget toward molecules that benefit from iterative refinement. The scorer measures embedding similarity rather than exact match, creating a reward-metric gap where structurally similar but incorrect molecules (e.g., isomers sharing NMR signatures) receive high rewards.

\subsection{Agentic Workflow Optimization}
\label{sec:exp-workflow}

\begin{table}[t]
\caption{Workflow optimization on DROP (F1 $\times$ 100, 3 seeds). $^*$Pilot-adapted (Section~\ref{sec:exploit-explore}).}
\label{tab:workflow-main}
\centering
\small
\begin{tabular}{lccc}
\toprule
\textbf{Method} & \textbf{F1 (mean$\pm$std)} & \textbf{Best} & \textbf{Cost (\$)} \\
\midrule
CoT (linear) & 89.28\tiny{$\pm$1.78} & 91.10 & 36.70 \\
AFlow & 87.89\tiny{$\pm$2.29} & 89.56 & 26.09 \\
\ours{} & \textbf{90.96}\tiny{$\pm$\underline{0.93}} & \textbf{91.54} & \textbf{15.26} \\
\ours{}$^*$ & \textbf{91.54}\tiny{$\pm$1.92} & \textbf{93.76} & \textbf{10.43} \\
\bottomrule
\end{tabular}
\end{table}

We evaluate \ours{} on DROP \citep{dua2019drop} within the AFlow framework \citep{aflow}, which searches over LLM-based operator graphs (20 rounds, 3 seeds, Claude Sonnet 4.5 optimizer, Haiku 4.5 executor). The three methods differ in how they condition refinement: CoT always extends the deepest chain; AFlow conditions on a top-scoring round chosen by stochastic sampling; \ours{} conditions on a tree-search-selected node.

CoT achieves strong accuracy by accumulating history but incurs the highest cost as context grows linearly. AFlow reduces cost by avoiding deep chains but sacrifices robustness through stateless selection. \ours{} is Pareto-dominant: tree search identifies productive nodes at moderate depths while quality gates prevent overcommitment to deep chains, achieving the best accuracy, lowest variance, and lowest cost (Table~\ref{tab:workflow-main}). We report cost only for this domain because the optimizer receives the parent workflow as context, making cost proportional to conditioning depth; in other domains (code generation, molecular elucidation), cost per iteration is fixed regardless of search depth.

% ============================================================================
% 6. ANALYSIS
% ============================================================================
\section{Understanding \ours{}}

\subsection{The Exploration Constant Is Not Enough}
\label{sec:exploration-constant}

A natural question is whether simply tuning the exploration constant $C$ in standard MCTS could replicate \ours{}'s gains. On LiveCodeBench (Figure~\ref{fig:code-main}), sweeping $C$ from 1.0 to 2.0 produces only marginal differences (44.3--45.6\% overall), with no configuration approaching \ours{} (46.2\%). The exploration-exploitation ratio is a single scalar that uniformly scales the bonus across all nodes; it cannot address flat reward signals, forced full expansion, or score-agnostic branching. Lowering $C$ shifts budget toward exploitation but still expands unpromising nodes unconditionally; raising $C$ broadens coverage but wastes budget on shallow siblings that never deepen. Neither direction addresses the core issue: the tree policy lacks the structural mechanisms to decide \emph{where} expansion is worthwhile. \ours{}'s improvement is structural: it reshapes \emph{how} the tree grows, not merely \emph{how much} it explores.

\subsection{Pilot-Run Findings}
\label{sec:exploit-explore}

We examine how pilot-run diagnostics (Table~\ref{tab:domains}) characterize the structural differences among agentic search problems and relate these differences to hyperparameter sensitivity. The high cost of agentic search experiments does not permit exhaustive grid search, so the \ours{}$^*$ configurations below may not represent optimal settings. Nonetheless, even coarse adaptation guided by these diagnostics yields \ours{}$^*$ variants that outperform all task-specific baselines (\rev{Tables~\ref{tab:prompt-main}, \ref{tab:mol-main}, \ref{tab:workflow-main} and Figure~\ref{fig:code-main}}). \rev{Appendix~\ref{app:diagnostic-guide} condenses these findings into a compact diagnostic guide (Table~\ref{tab:diagnostic-guide}).} Three patterns emerge.

\paragraph{Low $\hat{\sigma}_{\mathcal{A}}$/low $\kappa$ or unstable validation $\rightarrow$ soften $T$.} When refinement variance and drift are both low, scores cluster tightly and aggressive exploitation (low $T$) over-commits to gaps that may not predict test improvement. Relaxing $T$ to 0.5 hedges against this. On K-MSE ($\hat{\sigma}_{\mathcal{A}}\!=\!0.09$, $\kappa\!=\!0.05$), $T\!=\!0.5$ yields 91.2\% ACC ($+$0.2\,pp; Table~\ref{tab:mol-main}). On DROP ($\hat{\sigma}_{\mathcal{A}}\!=\!0.04$, $\kappa\!=\!0.02$), raising $\tau\!=\!0.5$ with $M_0\!=\!4$ yields 91.54 F1 ($+$0.58\,pp; Table~\ref{tab:workflow-main}). LiveCodeBench ($\hat{\sigma}_{\mathcal{A}}\!=\!0.32$, $\kappa\!=\!0.14$) does not exhibit this signature, yet also benefits from $T\!=\!0.5$ (47.4\%, $+$1.2\,pp; Figure~\ref{fig:code-main}) because its validation signal is inherently sparse: a few public test cases serve as proxy for the full hidden suite, creating per-sample noise that similarly rewards softer exploitation. In all three cases, moderate $T$-softening improves performance when the validation scorer is a weak proxy for the true metric.

\paragraph{High score drift ($\kappa$) $\rightarrow$ early exploration.} High $\kappa$ means normalization bounds shift frequently, making early shaped reward rankings unreliable. UCT's exploration term is larger than PUCT's when the tree is small, providing stronger early exploration before rankings stabilize. Disabling the score gate ($\tau_\text{gate}\!=\!0$) avoids blocking potentially good parents based on transiently low scores. On HoVeR ($\kappa = 0.16$, highest), switching to UCT with $\tau_\text{gate}\!=\!0$ yields 51.67\% ($+$1.22\,pp, $2.2\times$ lower variance; Table~\ref{tab:prompt-main}). Conversely, HotpotQA ($\kappa = 0.11$) strongly prefers PUCT ($+$6.45\,pp over UCT; Table~\ref{tab:ablation-exploration}), where fast decay concentrates budget on chains whose rankings are stable enough to trust.

\paragraph{High failure rate ($\rho$) $\rightarrow$ sensitive to $\alpha$.} Intuitively, when most expansions fail, the success-rate exponent $\alpha$ becomes critical for separating productive nodes from unproductive ones. We verify this on HotpotQA ($\rho = 0.74$): setting $\alpha\!=\!2.0$ yields 66.00\% ($+$1.11\,pp; Table~\ref{tab:prompt-main}), while reducing $\alpha$ to 1.0 causes $-$7.78\,pp with $6.8\times$ higher variance (Table~\ref{tab:ablation-components}). In low-failure tasks, $\alpha$ has negligible effect.

\subsection{Ablation Studies}
\label{sec:ablations}

We isolate each design choice via leave-one-out ablation on HotpotQA (high $\rho$, high $\kappa$). LiveCodeBench ablations are in Table~\ref{tab:ablation-full} (Appendix~\ref{app:ablation-full}).

\newcommand{\ablbar}[1]{\rule{3pt}{#1em}}
\begin{table}[t]
\centering
\small
\caption{Leave-one-out ablation on HotpotQA (3 seeds). LiveCodeBench ablations in Table~\ref{tab:ablation-full}.}
\label{tab:ablation-components}
\resizebox{\columnwidth}{!}{%
\begin{tabular}{lc@{\hspace{3pt}}c@{\hspace{12pt}}lc@{\hspace{3pt}}c}
\toprule
\textbf{Configuration} & \textbf{Acc.} & & \textbf{Configuration} & \textbf{Acc.} & \\
\midrule
\ours{} (full) & \textbf{64.89}\tiny{$\pm$0.83} & \ablbar{0.80} & w/o Soft norm. & 63.44\tiny{$\pm$2.20} & \ablbar{0.67} \\
w/o Virtual child & 61.67\tiny{$\pm$1.96} & \ablbar{0.51} & w/o Gated expan. & 63.22\tiny{$\pm$1.10} & \ablbar{0.65} \\
w/o Fail.\ penalty ($\alpha\!=\!1$) & 57.11\tiny{$\pm$5.67} & \ablbar{0.10} & w/o Gated PW & 62.55\tiny{$\pm$2.08} & \ablbar{0.59} \\
\bottomrule
\end{tabular}%
}
\end{table}

Table~\ref{tab:ablation-components} confirms that all components contribute. Failure penalty has the largest impact ($-$7.78\,pp), followed by the virtual child ($-$3.22\,pp); without the latter, the algorithm reverts to wide-tree behavior. Gated progressive widening contributes $-$2.34\,pp by restricting expansion to nodes whose quality justifies the budget cost.

\paragraph{Classical mechanisms do not transfer directly.}
Table~\ref{tab:ablation-classical} replaces \ours{} components with their classical counterparts: fixed FPU \citep{gelly2006fpu} instead of the virtual child, and score-agnostic progressive widening \citep{coulom2007computing} instead of quality-gated widening. On DROP, both FPU constants underperform \ours{} ($-$2.0\,pp and $-$1.6\,pp), and the gap between FPU$\,{=}\,0.5$ and FPU$\,{=}\,0.8$ ($+$0.4\,pp) illustrates a fundamental limitation: fixed FPU is sensitive to the constant's value, which interacts with exploration $C$, shaping temperature $T$, and domain score scale. The virtual child sidesteps this coupling by sampling from observed rewards, adapting automatically as the score distribution evolves. Score-agnostic widening wastes budget expanding low-quality nodes.

\begin{table}[t]
\centering
\small
\caption{Classical mechanisms vs.\ \ours{} redesigns (3 seeds).}
\label{tab:ablation-classical}
\begin{tabular}{lc}
\toprule
\textbf{Configuration} & \textbf{Score} \\
\midrule
\multicolumn{2}{l}{\textit{DROP (F1 $\times$ 100)}} \\
\quad \ours{} (virtual child) & \textbf{90.96}\tiny{$\pm$0.93} \\
\quad Fixed FPU $= 0.5$ & 88.95\tiny{$\pm$0.81} \\
\quad Fixed FPU $= 0.8$ & 89.34\tiny{$\pm$3.23} \\
\addlinespace
\multicolumn{2}{l}{\textit{LiveCodeBench (pass@1 \%)}} \\
\quad \ours{} (quality-gated PW) & \textbf{46.2}\tiny{$\pm$0.4} \\
\quad Score-agnostic PW & 45.0\tiny{$\pm$0.9} \\
\bottomrule
\end{tabular}
\end{table}

\paragraph{Exploration type: PUCT vs.\ UCT.}
Table~\ref{tab:ablation-exploration} compares the default PUCT (linear decay) against UCT (logarithmic decay). HotpotQA strongly prefers PUCT ($+$6.45\,pp), where fast decay concentrates budget on proven chains. HoVeR prefers UCT ($+$0.88\,pp), where sustained exploration buffers against score drift. This task-dependent sensitivity motivates the diagnostic analysis in Section~\ref{sec:exploit-explore}.

\begin{table}[t]
\centering
\small
\caption{Exploration type ablation: PUCT (default) vs.\ UCT, 3 seeds.}
\label{tab:ablation-exploration}
\begin{tabular}{lcc}
\toprule
\textbf{Dataset} & \textbf{PUCT} & \textbf{UCT} \\
\midrule
HotpotQA & \textbf{64.89}\tiny{$\pm$0.83} & 58.44\tiny{$\pm$2.20} \\
HoVeR & 50.45\tiny{$\pm$2.18} & \textbf{51.33}\tiny{$\pm$1.44} \\
K-MSE & \textbf{91.0}\tiny{$\pm$0.6} & 89.0\tiny{$\pm$1.5} \\
\bottomrule
\end{tabular}
\end{table}

\subsection{Tree-Shape Diagnostics}
\label{sec:tree-shape}

We verify that \ours{} redirects budget from breadth-first exploration into selective refinement by comparing tree shape on K-MSE (216 molecules, $B\!=\!16$, 3 seeds), where MCTSr and \ours{} differ only in tree policy. We measure mean leaf depth and depth of the best-scoring node, partitioning molecules into \emph{trivial} (124/216 solved at root) and \emph{non-trivial} (92/216). Full setup details are in Appendix~\ref{app:tree-shape-setup}.

\begin{table}[t]
\centering
\small
\caption{Tree-shape diagnostics on K-MSE ($B\!=\!16$, 3 seeds). Non-trivial: 92 molecules where at least one method improves beyond the root.}
\label{tab:tree-shape}
\resizebox{\columnwidth}{!}{%
\begin{tabular}{l cccc}
\toprule
\textbf{Method} & Leaf dep. & Best dep. & Scorer & ACC (\%) \\
\midrule
\multicolumn{5}{l}{\emph{All tasks} ($n\!=\!216$)} \\
\addlinespace[2pt]
MCTSr & 4.89\tiny{$\pm$.08} & 0.44\tiny{$\pm$.07} & 72.2\tiny{$\pm$0.3} & 89.8\tiny{$\pm$0.4} \\
\ours{}$_{M_0=2}$ & 5.49\tiny{$\pm$.13} & 0.48\tiny{$\pm$.05} & 72.5\tiny{$\pm$0.2} & 90.4\tiny{$\pm$0.4} \\
\ours{}$_{M_0=3}$ & 4.91\tiny{$\pm$.14} & 0.47\tiny{$\pm$.03} & \textbf{72.7}\tiny{$\pm$0.1} & \textbf{91.0}\tiny{$\pm$0.6} \\
\addlinespace[4pt]
\multicolumn{5}{l}{\emph{Non-trivial tasks} ($n\!=\!92$)} \\
\addlinespace[2pt]
MCTSr & 4.43\tiny{$\pm$.13} & 1.03\tiny{$\pm$.16} & 69.5\tiny{$\pm$0.6} & 77.2\tiny{$\pm$0.9} \\
\ours{}$_{M_0=2}$ & 4.44\tiny{$\pm$.09} & 1.12\tiny{$\pm$.11} & 69.6\tiny{$\pm$0.4} & 78.6\tiny{$\pm$1.0} \\
\ours{}$_{M_0=3}$ & 3.73\tiny{$\pm$.22} & 1.09\tiny{$\pm$.06} & \textbf{70.1}\tiny{$\pm$0.1} & \textbf{80.1}\tiny{$\pm$1.4} \\
\bottomrule
\end{tabular}%
}
\end{table}

On the non-trivial partition, \ours{} ($M_0\!=\!3$) produces shallower leaf depths (3.73 vs.\ 4.43) while finding its best node \emph{deeper} (1.09 vs.\ 1.03), the signature of selective refinement. This structural shift yields $+$2.9\,pp ACC on the non-trivial subset (80.1\% vs.\ 77.2\%), where the overall $+$1.2\,pp gain is concentrated. MCTSr distributes budget uniformly across depths; \ours{} concentrates children at promising nodes and deepens only subtrees that improve, placing the best node at the depth where refinement transitions from improving to diminishing returns.

\subsection{Budget-Scaling Behavior}
\label{sec:budget-scaling}

We measure performance as a function of evaluation budget $B$ across LiveCodeBench, HotpotQA, and K-MSE (setup details in Appendix~\ref{app:budget-scaling-setup}).

\begin{figure}[t]
\centering
\includegraphics[width=\columnwidth]{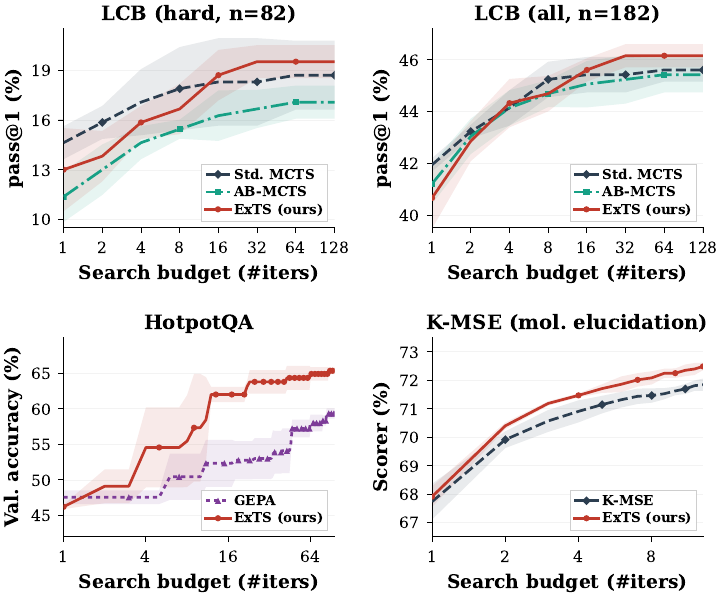}
\caption{Budget-scaling behavior across three domains, averaged over 3 seeds ($\pm1\sigma$ bands). The x-axis is search budget. \ours{}'s advantage is most pronounced on hard instances (top left) and in prompt optimization (bottom left), where iterative refinement matters most.}
\label{fig:budget-scaling}
\end{figure}

Figure~\ref{fig:budget-scaling} shows that \ours{} converts budget into performance more efficiently across all domains. On LiveCodeBench, \ours{} trails at $B{=}1$ (all methods reduce to a single sample) but overtakes Standard MCTS at $B{\approx}8$ and reaches 46.2\% vs.\ 45.6\% (Standard) and 45.4\% (AB-MCTS) at $B{=}128$; on hard problems the gap widens to \rev{19.1}\% vs.\ 17.1\% (AB-MCTS). On HotpotQA, \ours{} reaches 64.9\% at $B{=}70$ versus 58.0\% for GEPA Pareto with tighter confidence intervals. On K-MSE, \ours{} leads from $B{=}1$ onward and maintains a growing gap (72.5 vs.\ 71.8 at $B{=}13$). The crossover pattern reflects \ours{}'s calibration overhead: progressive widening and virtual-child scoring require initial samples, but this investment pays off once quality-conditioned pruning can redirect evaluations away from unpromising subtrees. The steeper scaling on hard problems ($+6.5$\,pp from $B{=}1$ to $B{=}32$ vs.\ $+3.7$\,pp for MCTS) confirms that the advantage concentrates where search depth matters most.

% ============================================================================
% 7. CONCLUSION
% ============================================================================
\section{Conclusion}

We introduced \ours{}, a tree-search policy for budget-constrained LLM agent search that jointly redesigns both selection and expansion. Rather than applying flat UCT selection and expanding unconditionally, \ours{} makes both decisions quality-aware: shaping rewards for narrow score distributions, framing expansion as a value-of-information decision, and gating branching on node quality. Across four diverse domains, \ours{} is competitive with or improves over task-specific baselines using a single fixed configuration, with gains concentrated in high-failure-rate and moderate-difficulty regimes. \rev{The same fixed configuration further generalizes to an additional domain, GPU kernel optimization (Appendix~\ref{app:gpu-kernel}).} We additionally show that pilot-run diagnostics characterize the structural differences among agentic search problems, explain why different tasks respond to different configurations, and offer practical guidance for adaptation.

% ============================================================================
% LIMITATIONS
% ============================================================================
\section*{Limitations}

We use a single model per domain due to the substantial API costs of tree search (each experiment requires hundreds of LLM calls per task instance across multiple seeds) and do not study how model scale alters the search landscape; because the pilot-run diagnostics are conditioned on the model, scorer, and budget, they describe the landscape \emph{as seen by that configuration} rather than intrinsic domain properties, and their transferability across model families is untested. Finally, the diagnostic approach requires a pilot tree that, in deployment on a new task, consumes budget not applied to the final search.

% ============================================================================
% REFERENCES
% ============================================================================
\bibliography{references}

@article{mlmaster,
  title={Ml-master: Towards ai-for-ai via integration of exploration and reasoning},
  author={Liu, Zexi and Cai, Yuzhu and Zhu, Xinyu and Zheng, Yujie and Chen, Runkun and Wen, Ying and Wang, Yanfeng and Chen, Siheng and others},
  journal={arXiv preprint arXiv:2506.16499},
  year={2025}
}

@article{mlzero,
  title={Mlzero: A multi-agent system for end-to-end machine learning automation},
  author={Fang, Haoyang and Han, Boran and Erickson, Nick and Zhang, Xiyuan and Zhou, Su and Dagar, Anirudh and Zhang, Jiani and Turkmen, Ali Caner and Hu, Tony and Rangwala, Huzefa and others},
  journal={Advances in Neural Information Processing Systems},
  volume={38},
  pages={69001--69070},
  year={2026}
}

@article{madaan2023selfrefine,
  title={Self-refine: Iterative refinement with self-feedback},
  author={Madaan, Aman and Tandon, Niket and Gupta, Prakhar and Hallinan, Skyler and Gao, Luyu and Wiegreffe, Sarah and Alon, Uri and Dziri, Nouha and Prabhumoye, Shrimai and Yang, Yiming and others},
  journal={Advances in neural information processing systems},
  volume={36},
  pages={46534--46594},
  year={2023}
}

@inproceedings{kmse,
  title={Boosting LLM’s molecular structure elucidation with knowledge enhanced tree search reasoning},
  author={Zhuang, Xiang and Wu, Bin and Cui, Jiyu and Feng, Kehua and Li, Xiaotong and Xing, Huabin and Ding, Keyan and Zhang, Qiang and Chen, Huajun},
  booktitle={Proceedings of the 63rd Annual Meeting of the Association for Computational Linguistics (Volume 1: Long Papers)},
  pages={22561--22576},
  year={2025}
}

@article{coulom2007computing,
  title={Computing “elo ratings” of move patterns in the game of go},
  author={Coulom, R{\'e}mi},
  journal={ICGA journal},
  volume={30},
  number={4},
  pages={198--208},
  year={2007},
  publisher={SAGE Publications Sage UK: London, England}
}

@inproceedings{kocsis2006uct,
  title={Bandit based monte-carlo planning},
  author={Kocsis, Levente and Szepesv{\'a}ri, Csaba},
  booktitle={European conference on machine learning},
  pages={282--293},
  year={2006},
  organization={Springer}
}

@article{browne2012mcts,
  title={A survey of monte carlo tree search methods},
  author={Browne, Cameron B and Powley, Edward and Whitehouse, Daniel and Lucas, Simon M and Cowling, Peter I and Rohlfshagen, Philipp and Tavener, Stephen and Perez, Diego and Samothrakis, Spyridon and Colton, Simon},
  journal={IEEE Transactions on Computational Intelligence and AI in games},
  volume={4},
  number={1},
  pages={1--43},
  year={2012},
  publisher={IEEE}
}

@article{agrawal2025gepa,
  title={Gepa: Reflective prompt evolution can outperform reinforcement learning},
  author={Agrawal, Lakshya A and Tan, Shangyin and Soylu, Dilara and Ziems, Noah and Khare, Rishi and Opsahl-Ong, Krista and Singhvi, Arnav and Shandilya, Herumb and Ryan, Michael J and Jiang, Meng and others},
  journal={arXiv preprint arXiv:2507.19457},
  year={2025}
}

@inproceedings{opsahlong2024miprov2,
  title={Optimizing instructions and demonstrations for multi-stage language model programs},
  author={Opsahl-Ong, Krista and Ryan, Michael J and Purtell, Josh and Broman, David and Potts, Christopher and Zaharia, Matei and Khattab, Omar},
  booktitle={Proceedings of the 2024 Conference on Empirical Methods in Natural Language Processing},
  pages={9340--9366},
  year={2024}
}

@article{yao2023tot,
  title={Tree of thoughts: Deliberate problem solving with large language models},
  author={Yao, Shunyu and Yu, Dian and Zhao, Jeffrey and Shafran, Izhak and Griffiths, Tom and Cao, Yuan and Narasimhan, Karthik},
  journal={Advances in neural information processing systems},
  volume={36},
  pages={11809--11822},
  year={2023}
}

@article{shinn2023reflexion,
  title={Reflexion: Language agents with verbal reinforcement learning},
  author={Shinn, Noah and Cassano, Federico and Gopinath, Ashwin and Narasimhan, Karthik and Yao, Shunyu},
  journal={Advances in neural information processing systems},
  volume={36},
  pages={8634--8652},
  year={2023}
}

@article{zhou2024lats,
  title={Language agent tree search unifies reasoning acting and planning in language models},
  author={Zhou, Andy and Yan, Kai and Shlapentokh-Rothman, Michal and Wang, Haohan and Wang, Yu-Xiong},
  journal={arXiv preprint arXiv:2310.04406},
  year={2023}
}

@article{khattab2024dspy,
  title={Dspy: Compiling declarative language model calls into self-improving pipelines},
  author={Khattab, Omar and Singhvi, Arnav and Maheshwari, Paridhi and Zhang, Zhiyuan and Santhanam, Keshav and Vardhamanan, Sri and Haq, Saiful and Sharma, Ashutosh and Joshi, Thomas T and Moazam, Hanna and others},
  journal={arXiv preprint arXiv:2310.03714},
  year={2023}
}

@inproceedings{yang2018hotpotqa,
  title={HotpotQA: A dataset for diverse, explainable multi-hop question answering},
  author={Yang, Zhilin and Qi, Peng and Zhang, Saizheng and Bengio, Yoshua and Cohen, William and Salakhutdinov, Ruslan and Manning, Christopher D},
  booktitle={Proceedings of the 2018 conference on empirical methods in natural language processing},
  pages={2369--2380},
  year={2018}
}

@inproceedings{jiang2020hover,
  title={HoVer: A dataset for many-hop fact extraction and claim verification},
  author={Jiang, Yichen and Bordia, Shikha and Zhong, Zheng and Dognin, Charles and Singh, Maneesh and Bansal, Mohit},
  booktitle={Findings of the Association for Computational Linguistics: EMNLP 2020},
  pages={3441--3460},
  year={2020}
}

@misc{jain2024livecodebench,
      title={LiveCodeBench: Holistic and Contamination Free Evaluation of Large Language Models for Code}, 
      author={Naman Jain and King Han and Alex Gu and Wen-Ding Li and Fanjia Yan and Tianjun Zhang and Sida Wang and Armando Solar-Lezama and Koushik Sen and Ion Stoica},
      year={2024},
      eprint={2403.07974},
      archivePrefix={arXiv},
      primaryClass={cs.SE},
      url={https://arxiv.org/abs/2403.07974}, 
}

@article{inoue2025abmcts,
  title={Wider or deeper? scaling llm inference-time compute with adaptive branching tree search},
  author={Inoue, Yuichi and Misaki, Kou and Imajuku, Yuki and Kuroki, So and Nakamura, Taishi and Akiba, Takuya},
  journal={Advances in Neural Information Processing Systems},
  volume={38},
  pages={35448--35484},
  year={2026}
}

@article{qwen2025qwen3,
  title={Qwen3 technical report},
  author={Yang, An and Li, Anfeng and Yang, Baosong and Zhang, Beichen and Hui, Binyuan and Zheng, Bo and Yu, Bowen and Gao, Chang and Huang, Chengen and Lv, Chenxu and others},
  journal={arXiv preprint arXiv:2505.09388},
  year={2025}
}

@inproceedings{santhanam2022colbertv2,
  title={Colbertv2: Effective and efficient retrieval via lightweight late interaction},
  author={Santhanam, Keshav and Khattab, Omar and Saad-Falcon, Jon and Potts, Christopher and Zaharia, Matei},
  booktitle={Proceedings of the 2022 Conference of the North American Chapter of the Association for Computational Linguistics: Human Language Technologies},
  pages={3715--3734},
  year={2022}
}

@article{yao2023react,
  title={React: Synergizing reasoning and acting in language models},
  author={Yao, Shunyu and Zhao, Jeffrey and Yu, Dian and Du, Nan and Shafran, Izhak and Narasimhan, Karthik and Cao, Yuan},
  journal={arXiv preprint arXiv:2210.03629},
  year={2022}
}

@article{schick2023toolformer,
  title={Toolformer: Language models can teach themselves to use tools},
  author={Schick, Timo and Dwivedi-Yu, Jane and Dess{\`\i}, Roberto and Raileanu, Roberta and Lomeli, Maria and Hambro, Eric and Zettlemoyer, Luke and Cancedda, Nicola and Scialom, Thomas},
  journal={Advances in neural information processing systems},
  volume={36},
  pages={68539--68551},
  year={2023}
}

@article{zhang2023planning,
  title={Planning with large language models for code generation},
  author={Zhang, Shun and Chen, Zhenfang and Shen, Yikang and Ding, Mingyu and Tenenbaum, Joshua B and Gan, Chuang},
  journal={arXiv preprint arXiv:2303.05510},
  year={2023}
}

@inproceedings{besta2024got,
  title={Graph of thoughts: Solving elaborate problems with large language models},
  author={Besta, Maciej and Blach, Nils and Kubicek, Ales and Gerstenberger, Robert and Podstawski, Michal and Gianinazzi, Lukas and Gajda, Joanna and Lehmann, Tomasz and Niewiadomski, Hubert and Nyczyk, Piotr and others},
  booktitle={Proceedings of the AAAI conference on artificial intelligence},
  volume={38},
  number={16},
  pages={17682--17690},
  year={2024}
}

@article{silver2016alphago,
  title={Mastering the game of Go with deep neural networks and tree search},
  author={Silver, David and Huang, Aja and Maddison, Chris J and Guez, Arthur and Sifre, Laurent and Van Den Driessche, George and Schrittwieser, Julian and Antonoglou, Ioannis and Panneershelvam, Veda and Lanctot, Marc and others},
  journal={nature},
  volume={529},
  number={7587},
  pages={484--489},
  year={2016},
  publisher={Nature Publishing Group UK London}
}

@article{rosin2011puct,
  title={Multi-armed bandits with episode context},
  author={Rosin, Christopher D},
  journal={Annals of Mathematics and Artificial Intelligence},
  volume={61},
  number={3},
  pages={203--230},
  year={2011},
  publisher={Springer}
}

@article{silver2017alphazero,
  title={Mastering chess and shogi by self-play with a general reinforcement learning algorithm},
  author={Silver, David and Hubert, Thomas and Schrittwieser, Julian and Antonoglou, Ioannis and Lai, Matthew and Guez, Arthur and Lanctot, Marc and Sifre, Laurent and Kumaran, Dharshan and Graepel, Thore and others},
  journal={arXiv preprint arXiv:1712.01815},
  year={2017}
}

@inproceedings{yang2024opro,
  title={Large language models as optimizers},
  author={Yang, Chengrun and Wang, Xuezhi and Lu, Yifeng and Liu, Hanxiao and Le, Quoc V and Zhou, Denny and Chen, Xinyun},
  booktitle={International Conference on Learning Representations},
  volume={2024},
  pages={12028--12068},
  year={2024}
}

@inproceedings{wang2024promptagent,
  title={Promptagent: Strategic planning with language models enables expert-level prompt optimization},
  author={Wang, Xinyuan and Li, Chenxi and Wang, Zhen and Bai, Fan and Luo, Haotian and Zhang, Jiayou and Jojic, Nebojsa and Xing, Eric and Hu, Zhiting},
  booktitle={International Conference on Learning Representations},
  volume={2024},
  pages={23967--24001},
  year={2024}
}

@inproceedings{gelly2006fpu,
  title={Exploration exploitation in go: UCT for Monte-Carlo go},
  author={Gelly, Sylvain and Wang, Yizao},
  booktitle={NIPS: Neural Information Processing Systems Conference On-line trading of Exploration and Exploitation Workshop},
  year={2006}
}

@article{chaslot2008progressive,
  title={Progressive strategies for Monte-Carlo tree search},
  author={Chaslot, Guillaume M Jb and Winands, Mark HM and Herik, H Jaap van den and Uiterwijk, Jos WHM and Bouzy, Bruno},
  journal={New Mathematics and Natural Computation},
  volume={4},
  number={03},
  pages={343--357},
  year={2008},
  publisher={World Scientific}
}

@inproceedings{aflow,
  title={Aflow: Automating agentic workflow generation},
  author={Zhang, Jiayi and Xiang, Jinyu and Yu, Zhaoyang and Teng, Fengwei and Chen, Xionghui and Chen, Jiaqi and Zhuge, Mingchen and Cheng, Xin and Hong, Sirui and Wang, Jinlin and others},
  booktitle={International Conference on Learning Representations},
  volume={2025},
  pages={34040--34077},
  year={2025}
}

@inproceedings{dua2019drop,
  title={DROP: A reading comprehension benchmark requiring discrete reasoning over paragraphs},
  author={Dua, Dheeru and Wang, Yizhong and Dasigi, Pradeep and Stanovsky, Gabriel and Singh, Sameer and Gardner, Matt},
  booktitle={Proceedings of the 2019 Conference of the North American Chapter of the Association for Computational Linguistics: Human Language Technologies, Volume 1 (Long and Short Papers)},
  pages={2368--2378},
  year={2019}
}

@misc{zhang2024llamaberry,
      title={LLaMA-Berry: Pairwise Optimization for O1-like Olympiad-Level Mathematical Reasoning}, 
      author={Di Zhang and Jianbo Wu and Jingdi Lei and Tong Che and Jiatong Li and Tong Xie and Xiaoshui Huang and Shufei Zhang and Marco Pavone and Yuqiang Li and Wanli Ouyang and Dongzhan Zhou},
      year={2024},
      eprint={2410.02884},
      archivePrefix={arXiv},
      primaryClass={cs.AI},
      url={https://arxiv.org/abs/2410.02884}, 
}

@misc{sstar,
      title={S*: Test Time Scaling for Code Generation}, 
      author={Dacheng Li and Shiyi Cao and Chengkun Cao and Xiuyu Li and Shangyin Tan and Kurt Keutzer and Jiarong Xing and Joseph E. Gonzalez and Ion Stoica},
      year={2025},
      eprint={2502.14382},
      archivePrefix={arXiv},
      primaryClass={cs.LG},
      url={https://arxiv.org/abs/2502.14382}, 
}

@misc{qi2024rstar,
      title={Mutual Reasoning Makes Smaller LLMs Stronger Problem-Solvers}, 
      author={Zhenting Qi and Mingyuan Ma and Jiahang Xu and Li Lyna Zhang and Fan Yang and Mao Yang},
      year={2024},
      eprint={2408.06195},
      archivePrefix={arXiv},
      primaryClass={cs.CL},
      url={https://arxiv.org/abs/2408.06195}, 
}

@misc{antoniades2024swesearch,
      title={SWE-Search: Enhancing Software Agents with Monte Carlo Tree Search and Iterative Refinement}, 
      author={Antonis Antoniades and Albert Örwall and Kexun Zhang and Yuxi Xie and Anirudh Goyal and William Wang},
      year={2025},
      eprint={2410.20285},
      archivePrefix={arXiv},
      primaryClass={cs.AI},
      url={https://arxiv.org/abs/2410.20285}, 
}

@misc{hao2023rap,
      title={Reasoning with Language Model is Planning with World Model}, 
      author={Shibo Hao and Yi Gu and Haodi Ma and Joshua Jiahua Hong and Zhen Wang and Daisy Zhe Wang and Zhiting Hu},
      year={2023},
      eprint={2305.14992},
      archivePrefix={arXiv},
      primaryClass={cs.CL},
      url={https://arxiv.org/abs/2305.14992}, 
}

@misc{li2026bavt,
      title={Spend Less, Reason Better: Budget-Aware Value Tree Search for LLM Agents},
      author={Yushu Li and Wenlong Deng and Jiajin Li and Xiaoxiao Li},
      year={2026},
      eprint={2603.12634},
      archivePrefix={arXiv},
      primaryClass={cs.LG},
      url={https://arxiv.org/abs/2603.12634},
}

@misc{snell2025scaling,
      title={Scaling LLM Test-Time Compute Optimally can be More Effective than Scaling Model Parameters}, 
      author={Charlie Snell and Jaehoon Lee and Kelvin Xu and Aviral Kumar},
      year={2024},
      eprint={2408.03314},
      archivePrefix={arXiv},
      primaryClass={cs.LG},
      url={https://arxiv.org/abs/2408.03314}, 
}

@misc{li2018hyperband,
      title={Hyperband: A Novel Bandit-Based Approach to Hyperparameter Optimization}, 
      author={Lisha Li and Kevin Jamieson and Giulia DeSalvo and Afshin Rostamizadeh and Ameet Talwalkar},
      year={2018},
      eprint={1603.06560},
      archivePrefix={arXiv},
      primaryClass={cs.LG},
      url={https://arxiv.org/abs/1603.06560}, 
}

@misc{brown2024bon,
      title={Large Language Monkeys: Scaling Inference Compute with Repeated Sampling},
      author={Bradley Brown and Jordan Juravsky and Ryan Ehrlich and Ronald Clark and Quoc V. Le and Christopher Ré and Azalia Mirhoseini},
      year={2024},
      eprint={2407.21787},
      archivePrefix={arXiv},
      primaryClass={cs.LG},
      url={https://arxiv.org/abs/2407.21787},
}

@misc{kwon2023vllm,
      title={Efficient Memory Management for Large Language Model Serving with PagedAttention}, 
      author={Woosuk Kwon and Zhuohan Li and Siyuan Zhuang and Ying Sheng and Lianmin Zheng and Cody Hao Yu and Joseph E. Gonzalez and Hao Zhang and Ion Stoica},
      year={2023},
      eprint={2309.06180},
      archivePrefix={arXiv},
      primaryClass={cs.LG},
      url={https://arxiv.org/abs/2309.06180}, 
}

@misc{wei2022cot,
      title={Chain-of-Thought Prompting Elicits Reasoning in Large Language Models}, 
      author={Jason Wei and Xuezhi Wang and Dale Schuurmans and Maarten Bosma and Brian Ichter and Fei Xia and Ed Chi and Quoc Le and Denny Zhou},
      year={2023},
      eprint={2201.11903},
      archivePrefix={arXiv},
      primaryClass={cs.CL},
      url={https://arxiv.org/abs/2201.11903}, 
}

@inproceedings{guan2024molpuzzle,
 author = {Guo, Kehan and Nan, Bozhao and Zhou, Yujun and Guo, Taicheng and Guo, Zhichun and Surve, Mihir and Liang, Zhenwen and Chawla, Nitesh V. and Wiest, Olaf and Zhang, Xiangliang},
 booktitle = {Advances in Neural Information Processing Systems},
 doi = {10.52202/079017-4281},
 editor = {A. Globerson and L. Mackey and D. Belgrave and A. Fan and U. Paquet and J. Tomczak and C. Zhang},
 pages = {134721--134746},
 publisher = {Curran Associates, Inc.},
 title = {Can LLMs Solve Molecule Puzzles? A Multimodal Benchmark for Molecular Structure Elucidation},
 url = {https://proceedings.neurips.cc/paper_files/paper/2024/file/f2b9e8e7a36d43ddfd3d55113d56b1e0-Paper-Datasets_and_Benchmarks_Track.pdf},
 volume = {37},
 year = {2024}
}

@inproceedings{shinka,
  title={Shinkaevolve: Towards open-ended and sample-efficient program evolution},
  author={Lange, Robert and Imajuku, Yuki and Cetin, Edoardo},
  booktitle={International Conference on Learning Representations},
  volume={2026},
  pages={74026--74078},
  year={2026}
}

@article{evox,
  title={Evox: Meta-evolution for automated discovery},
  author={Liu, Shu and Agarwal, Shubham and Maheswaran, Monishwaran and Cemri, Mert and Li, Zhifei and Mang, Qiuyang and Naren, Ashwin and Boneh, Ethan and Cheng, Audrey and Pan, Melissa Z and others},
  journal={arXiv preprint arXiv:2602.23413},
  year={2026}
}

@article{llmzero,
  title={LLMZero: Discovering Adaptive Training Strategies for RL Post-Training via LLM Agents},
  author={Fang, Haoyang and Zhu, Wei and Han, Boran and Zhang, Alex and Pan, Zhenyu and Yang, Shuo and Zhang, Shuai and Gai, Jiading and Tang, Peng and Hu, Cuixiong and others},
  journal={arXiv preprint arXiv:2606.18388},
  year={2026}
}

@article{ifco,
  title={Optimizing CUDA like a Human: Micro-Profiling Tools as Expert Surrogates for LLM-Based GPU Kernel Optimization},
  author={Gai, Jiading and Zhang, Shuai and Bostrom, Kaj and Huang, Jin and Patil, Vihang and Fang, Haoyang and Wang, Bernie and Rangwala, Huzefa and Karypis, George},
  journal={arXiv preprint arXiv:2606.26453},
  year={2026}
}

@article{reskill,
  title={ReSkill: Reconciling Skill Creation with Policy Optimization in Agentic RL},
  author={He, Zelin and Lin, Haotian and Han, Boran and Zhu, Wei and Fang, Haoyang and Wang, Bernie and Zhu, Xuan and Li, Runze and Reimherr, Matthew},
  journal={arXiv preprint arXiv:2606.01619},
  year={2026}
}

% ============================================================================
% APPENDIX
% ============================================================================
\newpage
\tableofcontents
\newpage
\appendix

\section{Related Work}
\label{sec:related}

\paragraph{Search for LLM reasoning and generation.}
Search strategies for LLM generation can be organized by whether their primary contribution lies in the \emph{harness} (the framework surrounding the search) or the \emph{search algorithm} itself (selection and expansion policies).

On the harness side, many systems wrap standard MCTS with LLM-specific components while retaining vanilla UCT for selection and expansion. Some operate as linear refinement chains without tree structure \citep{shinn2023reflexion, madaan2023selfrefine, mlzero, llmzero}. LATS \citep{zhou2024lats} integrates MCTS with LLM value functions and self-reflection but does not modify the tree policy. rStar \citep{qi2024rstar} defines rich reasoning actions with mutual verification between two SLMs. RAP \citep{hao2023rap} repurposes the LLM as a world model within standard MCTS. SWE-Search \citep{antoniades2024swesearch} introduces a multi-agent evaluation framework with a hybrid value function for software engineering tasks. PromptAgent \citep{wang2024promptagent} frames prompt optimization as MCTS-style strategic planning with error feedback. LLaMA-Berry \citep{zhang2024llamaberry} replaces raw value estimates with pairwise preference aggregation for mathematical reasoning but retains standard UCT selection. S* \citep{sstar} combines parallel sampling with sequential refinement and execution-grounded selection for code generation, contributing a scaling pipeline rather than a tree policy. MLMaster \citep{mlmaster} applies vanilla UCT to data science automation. In each case, the primary innovation is the surrounding framework or evaluation mechanism; the internal selection and expansion dynamics remain standard or unmodified, representing a gap where better search policies could yield further gains.

On the search algorithm side, fewer works redesign how budget is allocated across candidates. AB-MCTS \citep{inoue2025abmcts} replaces UCT with Thompson sampling for code generation. GEPA \citep{agrawal2025gepa} uses Pareto-frontier selection over scored candidates for prompt optimization. AFlow \citep{aflow} applies score-weighted random sampling from recent rounds for workflow optimization. BAVT \citep{li2026bavt} introduces budget-conditioned node selection via a power-law exponent that shifts from exploration to exploitation as budget depletes, evaluated on multi-hop QA tasks. BAVT is not open-sourced, precluding direct comparison.

{\color{revcolor}
\ours{} departs from most of these methods by making the exploration/exploitation balance \emph{reward-aware}: instead of a fixed exploration constant, a reward-agnostic first-play-urgency value, or progressive widening tied to visit count alone, it lets observed validation rewards drive both the value of expansion and how wide a node may branch. Its virtual child is the clearest example: where first-play urgency \citep{gelly2006fpu} assigns the same fixed value to every unexpanded child, the virtual child is a node-specific, data-dependent estimate resampled from the parent's shaped-reward pool and scored at a fair-share visit count, which we treat as a value-of-information-style estimate of expansion value. Its closest relative is the adaptive branching of AB-MCTS \citep{inoue2025abmcts} (one of our baselines), which likewise lets a hypothetical branch compete with existing children. However, AB-MCTS-A relies on parametric Bayesian posteriors with conjugate priors, while AB-MCTS-M dynamically estimates two posterior distributions via MCMC, which can be less effective when the search budget is small; the virtual child, by contrast, is a nonparametric, prior-free bootstrap coupled with quality-gated widening. We compare against AB-MCTS-A (Gaussian), the strongest AB-MCTS variant on LiveCodeBench. Finally, whereas each of these baselines targets a single task type, \ours{} redesigns both selection and expansion and holds a single fixed configuration across four structurally distinct domains, plugging into each framework's existing search interface (GEPA, TreeQuest, K-MSE, AFlow). Table~\ref{tab:positioning} summarizes this positioning against prior LLM tree-search methods.
}

\paragraph{Test-time compute allocation and sampling baselines.}
Several lines of work study inference-time budget allocation from complementary perspectives.
\citet{snell2025scaling} characterize when repeated sampling versus sequential revision is compute-optimal for reasoning tasks, providing scaling predictions at the problem level. \ours{} addresses a different use case, agentic search with iterative refinement, but its mechanisms for efficient budget allocation could in principle be applied within test-time scaling frameworks.
Best-of-$N$ sampling \citep{brown2024bon} generates $N$ independent candidates and returns the best. \ours{} instead exploits parent-child refinement structure, achieving better efficiency when iterative improvement is productive.
Sequential halving and Hyperband \citep{li2018hyperband} efficiently eliminate unpromising configurations but assume independent candidates. \ours{} handles tree-structured refinement where parent quality predicts child quality.

\begin{table}[h]
\caption{Positioning of \ours{} relative to prior LLM tree-search methods. ``Focus'' indicates whether the primary contribution is the search harness (H) or the search algorithm (A).}
\label{tab:positioning}
\centering
\small
\resizebox{\columnwidth}{!}{%
\begin{tabular}{lcccc}
\toprule
\textbf{Method} & \textbf{Focus} & \textbf{Selection} & \textbf{Expansion} & \textbf{Fail.-aware} \\
\midrule
Linear & H & Sequential & N/A & No \\
ToT & H & BFS/DFS & Full & No \\
LATS & H & UCT & Full & No \\
rStar & H & UCT & Full & No \\
RAP & H & UCT & Full & No \\
SWE-Search & H & UCT + hybrid value & Full & No \\
LLaMA-Berry & H & UCT + pairwise & Full & No \\
S* & H & Parallel + refine & Pipeline & No \\
MLMaster & H & UCT & Full & No \\
PromptAgent & H & UCT & Full & No \\
\midrule
GEPA & A & Pareto & Stateless & No \\
AFlow & A & Score-weighted & Random & No \\
AB-MCTS & A & Thompson & Adaptive & No \\
BAVT & A & Budget-conditioned & Structural & No \\
\textbf{\ours{}} & A & ExUCT & Gated + adaptive & Yes \\
\bottomrule
\end{tabular}%
}
\end{table}

\paragraph{\color{revcolor}Evolutionary search for LLM systems.}
{\color{revcolor}
ShinkaEvolve \citep{shinka} and EvoX \citep{evox} are population-based evolutionary optimizers that run at a far larger budget than \ours{}'s tight per-query regime (ShinkaEvolve uses roughly 150 evaluations for its headline result, whereas several of our settings run well below this, e.g., 16 rollouts on K-MSE and 20 rounds on AFlow), so they differ from \ours{} in search family, budget, and search space. Even GEPA \citep{agrawal2025gepa}, one of our prompt-optimization baselines, is itself a Genetic-Pareto evolutionary method, yet \ours{} improves over it by 10.8\% on HotpotQA at 6.4$\times$ lower variance. \ours{} is instead a complementary, plug-and-play tree policy whose value-of-information view of expansion is orthogonal to how candidates are generated, and matched-budget comparisons between tree and evolutionary search (and hybrids of the two) are a valuable direction for future work.
}

\paragraph{Prompt \rev{and Skill} optimization.}
OPRO \citep{yang2024opro} uses LLMs as optimizers over scored solution histories. MIPROv2 \citep{opsahlong2024miprov2} applies Bayesian surrogate optimization to DSPy instruction tuning. \rev{For skill optimization, ReSkill \citep{reskill} selects among candidate skills via Thompson sampling in agentic RL.} \ours{} can serve as a drop-in selection component within such frameworks.

\paragraph{Self-improving agents.}
Reflexion \citep{shinn2023reflexion} uses verbal self-reflection for iterative improvement but follows a linear chain rather than a tree, missing the opportunity to explore alternative refinement paths. ReAct \citep{yao2023react} synergizes reasoning and acting but uses greedy selection. These approaches are complementary, and \ours{} could serve as the search backbone for systems that currently rely on linear or greedy strategies.

\paragraph{Automated workflow optimization.}
AFlow \citep{aflow} automates agentic workflow generation by searching the space of LLM-based operator graphs. We directly compare against AFlow's search strategy in Section~\ref{sec:exp-workflow}.

\paragraph{Molecular structure elucidation.}
K-MSE \citep{kmse} already applies MCTS with a neural molecule-spectrum scorer to molecular structure elucidation, making it a directly relevant baseline. Our K-MSE evaluation uses the same scorer and knowledge base, isolating the effect of the selection/expansion policy.

\section{Default Configuration and Complete Algorithm}
\label{app:defaults}

Algorithm~\ref{alg:exts} gives the complete \ours{} search loop and Table~\ref{tab:defaults} lists the default hyperparameters used across all experiments.

\begin{algorithm}[h]
\caption{\ours{}: Complete Search Loop}
\label{alg:exts}
\begin{algorithmic}[1]
\REQUIRE $x_0$, refinement operator $\mathcal{A}$, validator $f$, budget $B$
\STATE Initialize tree $\mathcal{T}$ with root $v_0$, $x_{v_0} \leftarrow x_0$, $s_{v_0} \leftarrow f(x_0)$
\WHILE{budget $B$ not exhausted}
    \STATE $v \leftarrow \textsc{ExUCT-Select}(v_0)$ \hfill {\small Alg.~\ref{alg:uct-select}}
    \STATE $x' \leftarrow \mathcal{A}(x_v)$ \hfill {\small LLM refinement}
    \STATE $r \leftarrow f(x')$
    \IF{$r = \bot$}
        \STATE \textsc{Backprop-Failure}($v$);\, \textbf{continue}
    \ENDIF
    \STATE Create child $u$: $x_u \leftarrow x'$,\; $s_u \leftarrow r$
    \STATE $\mathrm{children}(v) \leftarrow \mathrm{children}(v) \cup \{u\}$
    \STATE \textsc{Backprop-Success}($v$, $r$)
\ENDWHILE
\RETURN $\arg\max_{v \in \mathcal{T}} s_v$
\end{algorithmic}
\end{algorithm}

\begin{table}[h]
\caption{\ours{} default configuration, fixed across all experiments.}
\label{tab:defaults}
\centering
\small
\resizebox{\columnwidth}{!}{%
\begin{tabular}{lcl}
\toprule
\textbf{Parameter} & \textbf{Default} & \textbf{Rationale} \\
\midrule
Exploration type & PUCT & Linear decay for tight budgets \\
Exploration $C$ & 1.0 & Conservative; no policy prior \\
Success-rate $\alpha$ & $\sqrt{2}$ & Amplified failure discounting \\
Temperature $T$ & 0.3 & Top-20\% emphasis \\
Initial children $M_0$ & 3 & Moderate branching \\
Widening base $b$ & 2 & $\log_2$ growth \\
Widening threshold $n_0$ & 32 & Delay widening \\
Widening quantile $\gamma$ & 0.75 & Top-quartile widening \\
Score gate $\tau$ & 0.25 & Mild gating \\
\bottomrule
\end{tabular}%
}
\end{table}

\section{Formal Definitions of Pilot-Run Diagnostics}
\label{app:properties}

Section~\ref{sec:validation-heavy} introduces four measurable pilot-run diagnostics that characterize the search landscape of validation-heavy domains. We provide rigorous definitions below, expressed in terms of the search tree $\mathcal{T}$ built during a pilot run of budget $B$. \rev{Table~\ref{tab:property-notation} summarizes the notation used throughout.}

\begin{table}[h]
\caption{Notation used in formal property definitions.}
\label{tab:property-notation}
\centering
\small
\resizebox{\columnwidth}{!}{%
\begin{tabular}{cl}
\toprule
\textbf{Symbol} & \textbf{Description} \\
\midrule
$\mathcal{T}$ & Search tree rooted at initial candidate $x_0$ \\
$V(\mathcal{T})$ & Set of all nodes in $\mathcal{T}$ \\
$I(\mathcal{T})$ & Set of internal (expanded) nodes: $\{v \in V(\mathcal{T}) : |\mathrm{children}(v)| > 0\}$ \\
$n_v$ & Visit count of node $v$ (incremented via backpropagation from all expansions in the subtree rooted at $v$) \\
$n_v^+$ & Successful visit count (backpropagated from expansions where $f(x') \neq \bot$) \\
$n_v^-$ & Failed visit count: $n_v^- = n_v - n_v^+$ \\
$s_v$ & Validation score $f(x_v)$ for node $v$'s candidate $x_v$ \\
$S^+$ & Set of successfully validated nodes: $\{v \in V(\mathcal{T}) : s_v \text{ is defined}\}$ \\
$s_{\min}, s_{\max}$ & $\min_{v \in S^+} s_v$ and $\max_{v \in S^+} s_v$ respectively \\
\bottomrule
\end{tabular}%
}
\end{table}

\paragraph{Definition 1: Failure rate ($\rho$).}
The fraction of expansion attempts that do not produce a successfully validated candidate:
\begin{equation}
\label{eq:failure-rate}
\rho \;=\; \frac{n_{v_0}^-}{n_{v_0}}
\end{equation}
where $v_0$ is the root. Since backpropagation increments all ancestors after each expansion, $n_{v_0}$ equals the total number of expansion attempts across the tree and $n_{v_0}^-$ equals the total failures. An attempt counts as failed if $f$ returns $\bot$ (e.g., syntax error, runtime error, timeout, or a domain-specific rejection criterion). By construction, $\rho \in [0, 1]$. Domains with $\rho > 0.5$ spend more than half their budget on failed attempts, making failure-aware mechanisms (dual backpropagation, success-rate weighting) essential.

\paragraph{Definition 2: Normalized score deviation ($\hat{\sigma}_f$).}
The spread of validation scores across all successfully evaluated candidates, normalized by the observed range:
\begin{equation}
\label{eq:score-deviation}
\hat{\sigma}_f \;=\; \frac{1}{s_{\max} - s_{\min}} \sqrt{\frac{1}{|S^+| - 1} \sum_{v \in S^+} (s_v - \bar{s})^2}
\end{equation}
where $\bar{s} = \frac{1}{|S^+|}\sum_{v \in S^+} s_v$. If $s_{\max} = s_{\min}$ or $|S^+| < 2$, we define $\hat{\sigma}_f = 0$. Low values indicate a flat score landscape where exploitation shaping must work harder to differentiate candidates.

\paragraph{Definition 3: Score drift ($\kappa$).}
Score drift quantifies the non-stationarity of normalized node scores as the tree grows. At each iteration $t$ a new node with score $s_{\mathrm{new}}$ is added. Let $s_{\min}^t$ and $s_{\max}^t$ denote the running minimum and maximum scores among all nodes at iteration $t$. Let $\mathcal{T}_{\Delta} = \{t : s_{\min}^t \neq s_{\min}^{t-1} \text{ or } s_{\max}^t \neq s_{\max}^{t-1}\}$ be iterations where the normalization bounds change. The normalized score of node $v$ at time $t$ is $\hat{s}_v^t = (s_v - s_{\min}^t)/(s_{\max}^t - s_{\min}^t)$. Then:
\begin{equation}
\label{eq:score-drift}
\kappa \;=\; \frac{1}{|\mathcal{T}_{\Delta}|} \sum_{t \in \mathcal{T}_{\Delta}} \frac{1}{|V_t|} \sum_{v \in V_t} \bigl|\hat{s}_v^t - \hat{s}_v^{t-1}\bigr|
\end{equation}
where $V_t$ is the set of nodes existing before iteration $t$ (excluding the newly added node). If $|\mathcal{T}_{\Delta}| = 0$ or the score range is zero, we define $\kappa = 0$. High $\kappa$ indicates that newly discovered candidates frequently shift the normalization bounds, changing the relative exploitation values of existing candidates.

\paragraph{Definition 4: Refinement variance ($\hat{\sigma}_{\mathcal{A}}$).}
The refinement variance measures the stochasticity of the refinement operator $\mathcal{A}$ when applied to the same parent state. For each internal node $v \in I(\mathcal{T})$ with at least two successfully validated children, let $C_v^+ = \{c \in \mathrm{children}(v) : s_c \text{ is defined}\}$ and $\bar{f}_v = \frac{1}{|C_v^+|}\sum_{c \in C_v^+} s_c$. The per-node refinement variance is:
\begin{equation}
\label{eq:per-node-refvar}
\sigma^2_{\mathcal{A},v} \;=\; \frac{1}{|C_v^+| - 1} \sum_{c \in C_v^+} (s_c - \bar{f}_v)^2
\end{equation}
Let $I_2 = \{v \in I(\mathcal{T}) : |C_v^+| \geq 2\}$. The global refinement variance is:
\begin{equation}
\label{eq:refinement-variance}
\hat{\sigma}_{\mathcal{A}} \;=\; \frac{1}{s_{\max} - s_{\min}} \sqrt{\frac{1}{|I_2|} \sum_{v \in I_2} \sigma^2_{\mathcal{A},v}}
\end{equation}

\paragraph{Distinction from $\hat{\sigma}_f$.}
$\hat{\sigma}_f$ measures the spread of scores across \emph{all} candidates in the tree, reflecting the combined effect of different parent states, depths, and refinement histories. In contrast, $\hat{\sigma}_{\mathcal{A}}$ isolates the variance attributable to the LLM's sampling stochasticity by conditioning on the parent. A domain may have low $\hat{\sigma}_f$ (flat overall landscape) but high $\hat{\sigma}_{\mathcal{A}}$ (diverse candidates from any single parent), or vice versa. High $\hat{\sigma}_{\mathcal{A}}$ indicates that re-sampling from the same node is productive, providing implicit exploration; low $\hat{\sigma}_{\mathcal{A}}$ indicates refinements are largely deterministic and explicit exploration via higher $C$ is needed.

\paragraph{Computation details for Table~\ref{tab:domains}.}
All properties are computed from pilot trees built by running the baseline method for each dataset with budget $B$, averaged across 3 seeds ($\{0, 42, 1024\}$).

\emph{$\rho$:} We read $n_{v_0}$ and $n_{v_0}^+$ directly from the root node's backpropagated statistics stored in the pilot tree.

\emph{$\hat{\sigma}_f$:} We collect the score $s_v$ of every node $v \in S^+$ in the pilot tree and compute the sample standard deviation divided by $s_{\max} - s_{\min}$. For K-MSE, which runs 216 independent molecule searches per seed, we pool all node scores across molecules before computing a single $\hat{\sigma}_f$ per seed.

\emph{$\kappa$:} We replay the tree construction in node-creation order, tracking $s_{\min}^t$ and $s_{\max}^t$ after each insertion. At each iteration where bounds change, we compute the mean absolute shift in normalized scores across existing nodes and average over all such events.

\emph{$\hat{\sigma}_{\mathcal{A}}$:} We estimate refinement variance from sibling scores in the pilot tree: for each internal node $v$ with $|C_v^+| \geq 2$ successfully validated children, we compute $\sigma^2_{\mathcal{A},v}$ from the children's scores and aggregate via Equation~\ref{eq:refinement-variance}. This treats siblings as approximate samples from $\mathcal{A}(x_v)$, which is valid when the tree context does not meaningfully change between sibling expansions. For LiveCodeBench, where the pilot does not save per-node tree structure, we instead run $K\!=\!10$ i.i.d.\ refinements from a fixed parent state for 5 representative problems and aggregate.

\section{Detailed Experimental Setup}
\label{app:setup}

\subsection{Prompt Optimization}
\label{app:setup-prompt}

We evaluate \ours{} within the GEPA prompt optimization framework \citep{agrawal2025gepa}, which optimizes natural-language instructions in DSPy programs \citep{khattab2024dspy}. We use two multi-hop reasoning benchmarks: HotpotQA \citep{yang2018hotpotqa}, a question answering task with a 4-predictor DSPy program and ColBERTv2 retrieval \citep{santhanam2022colbertv2} over 5.2M Wikipedia abstracts, and HoVeR \citep{jiang2020hover}, a fact verification task. Both use a test set of 300 examples with accuracy (\%) as the evaluation metric, and all experiments are run across seeds $\{0, 42, 1024\}$ using Qwen3-8B \citep{qwen2025qwen3} via vLLM \citep{kwon2023vllm} (TP=4, temperature 0.6, top-$p$ 0.95, thinking enabled).

In the \ours{} formulation, the candidate set $\mathcal{S}$ consists of DSPy instruction configurations. The refinement operator $\mathcal{A}$ is an LLM instruction proposer conditioned on failure feedback. The validator $f$ evaluates on the full train and validation set, returning $\bot$ if the new score does not exceed the parent's. Score stationarity $\kappa$ is high because aggregate scores shift when new programs join the Pareto frontier.

We compare four conditions: Baseline (unoptimized seed program), GEPA Pareto (task-native Pareto-frontier selection \citep{agrawal2025gepa}), \ours{} default (Table~\ref{tab:defaults}), and per-dataset \ours{}$^*$ (HotpotQA: $\alpha\!=\!2.0$; HoVeR: UCT exploration, $\tau\!=\!0$), with the optimization budget matched across methods.

\subsection{Code Generation}
\label{app:setup-code}

We evaluate \ours{} within the TreeQuest framework \citep{inoue2025abmcts} on LiveCodeBench \citep{jain2024livecodebench}, which contains 182 problems from release v6 (January to April 2025) split into 45 easy, 55 medium, and 82 hard problems. Public test cases serve as the search reward signal and private test cases determine the final pass@1 metric. We use Claude Sonnet 4 with temperature 0.6, a budget of 128 steps per problem, and report results over seeds $\{0, 42, 1024\}$.

In the \ours{} formulation, $\mathcal{S}$ consists of code solutions, $\mathcal{A}$ is an LLM code editor conditioned on test failure traces, and $f$ is the full public test suite execution (returning $\bot$ on syntax errors or compilation failures). Score stationarity $\kappa$ is low because existing solution scores do not change. We compare StandardMCTS (UCT with $C\!=\!\sqrt{2}$, samples per action of 5), AB-MCTS-A (Thompson sampling with Gaussian conjugate prior \citep{inoue2025abmcts}), and \ours{} default (Table~\ref{tab:defaults}). \ours{}$^*$ sets $T\!=\!0.5$.

\subsection{Molecular Structure Elucidation}
\label{app:setup-mol}

We evaluate \ours{} on the K-MSE molecular structure elucidation task \citep{kmse}, in which an LLM must deduce a molecule's SMILES representation from infrared (IR) and nuclear magnetic resonance (NMR) spectral data together with a molecular formula. The evaluation set consists of 216 molecules from the MolPuzzle dataset \citep{guan2024molpuzzle}, each paired with its IR spectrum image, carbon-13 and proton NMR spectra, and molecular formula. The primary metrics are Morgan fingerprint Tanimoto similarity (FTS) and exact-match accuracy (ACC), where ACC canonicalizes both predicted and ground-truth SMILES via RDKit before comparison. We use Claude Sonnet 4.6 with a pre-trained molecule-spectrum alignment scorer for reward computation.

In the \ours{} formulation, $\mathcal{S}$ consists of SMILES strings, $\mathcal{A}$ is an LLM that critiques the current prediction against spectral data and a retrieved knowledge base of 593 molecular substructures and then proposes a revised SMILES, $g$ checks chemical validity via RDKit, and $f$ is a neural scorer that computes cosine similarity between the predicted molecule's embedding and the target spectrum's embedding scaled to $[0, 100]$. We compare \ours{} against K-MSE's original MCTSr \citep{kmse} and chain-of-thought prompting \citep[CoT;][]{wei2022cot}, using 16 rollouts per molecule and 3 seeds ($\{0, 42, 1024\}$). \ours{} uses the default configuration (Table~\ref{tab:defaults}). \ours{}$^*$ sets $T\!=\!0.5$.

\subsection{Agentic Workflow Optimization}
\label{app:setup-workflow}

We evaluate \ours{} within the AFlow framework \citep{aflow} on the DROP reading comprehension benchmark \citep{dua2019drop}. DROP consists of paragraphs with questions requiring discrete reasoning (counting, sorting, arithmetic) over text; the test set contains 800 problems scored by token-level F1.

In the AFlow formulation, a ``workflow'' is a Python function that orchestrates one or more LLM calls (using operators such as Generate, Format, Review, Ensemble) to answer a question given a passage. The search space $\mathcal{S}$ consists of these workflow programs. The refinement operator $\mathcal{A}$ is an optimizer LLM (Claude Sonnet 4.5) that proposes mutations to the workflow graph conditioned on execution logs and past experience. The validator $f$ executes the workflow on a validation split of 200 problems over 5 rounds and returns the mean F1 score. The failure rate is low ($\rho \approx 0.05$) because workflows almost always produce parseable output; failures are rare execution errors.

We compare three search strategies sharing the same optimizer LLM, executor LLM (Claude Haiku 4.5), validation protocol, and test evaluation:
\begin{itemize}[nosep]
\item \textbf{AFlow} \citep{aflow}: Score-weighted random parent sampling from recent rounds. No tree structure or visit statistics.
\item \textbf{CoT (linear)}: Sequential refinement chain that always extends the deepest node. No branching or selection policy.
\item \textbf{\ours{}}: Default configuration (Table~\ref{tab:defaults}). \ours{}$^*$ sets $\tau\!=\!0.5$, $M_0\!=\!4$.
\end{itemize}
Budget is 20 search rounds per seed, with seeds $\{0, 42, 1024\}$.

\section{Analysis Setup Details}
\label{app:analysis-setup}

\paragraph{Tree-shape diagnostics (Section~\ref{sec:tree-shape}).}
\label{app:tree-shape-setup}
We compare MCTSr (binary expansion, no gating) against \ours{} at two branching caps: $M_0\!=\!2$ (matching MCTSr's effective branching) and $M_0\!=\!3$ (the default). All methods share the same scorer, retriever, and LLM (Claude Sonnet 4.6). Molecules are partitioned into \emph{trivial} (best node at depth~0 for all methods in all seeds; 124/216) and \emph{non-trivial} (at least one method in one seed improves beyond the root; 92/216). On the full dataset, all methods achieve similar leaf depths (${\sim}5$) because 57\% of molecules are trivially solved and MCTS exhausts budget on already-solved problems. The non-trivial partition isolates the algorithmic difference. The scorer gap on non-trivial molecules ($+$0.6 over MCTSr) translates to a meaningful ACC improvement because this partition selects precisely the molecules where the scorer provides actionable signal.

\paragraph{Budget-scaling behavior (Section~\ref{sec:budget-scaling}).}
\label{app:budget-scaling-setup}
On LiveCodeBench we run Standard MCTS, AB-MCTS, and \ours{} with $B{=}128$ (182 problems, 3 seeds) and extract pass@1 at each intermediate budget from the cumulative score trajectory: a problem is solved at budget $b$ if its best public-test solution within the first $b$ evaluations also passes all held-out private tests. On HotpotQA we compare \ours{} and GEPA Pareto over 78 search iterations (3 seeds), tracking cumulative best validation accuracy. On K-MSE we compare \ours{} and MCTSr over 13 evaluation steps (3 seeds, 216 molecules), tracking cumulative best scorer score.

\section{More Ablations on LiveCodeBench}
\label{app:ablation-full}

Table~\ref{tab:ablation-full} isolates two components on LiveCodeBench. Removing gated progressive widening costs $-$1.3\,pp overall and $-$1.6\,pp on hard problems, confirming that quality-conditioned branching prevents budget waste on unpromising subtrees. Removing root exclusion (allowing the score gate to block root expansion) costs $-$1.0\,pp overall and $-$2.0\,pp on hard, with notably higher variance, because gating the root can starve the tree of initial diversity when early candidates score poorly.

\begin{table}[h]
\centering
\small
\caption{Leave-one-out ablation on LiveCodeBench (pass@1 \%, 3 seeds).}
\label{tab:ablation-full}
\begin{tabular}{lcc}
\toprule
\textbf{Configuration} & \textbf{All (\%)} & \textbf{Hard (\%)} \\
\midrule
\ours{} (full) & \textbf{46.2}\tiny{$\pm$0.4} & \textbf{19.1}\tiny{$\pm$1.4} \\
\quad w/o Gated prog.\ widening & 44.9\tiny{$\pm$0.7} & 17.5\tiny{$\pm$1.4} \\
\quad w/o Root exclusion & 45.2\tiny{$\pm$1.5} & 17.1\tiny{$\pm$2.5} \\
\bottomrule
\end{tabular}
\end{table}

\section{{Compact Diagnostic Guide}}
\label{app:diagnostic-guide}

{\color{revcolor}
Table~\ref{tab:diagnostic-guide} condenses the pilot-run findings of Section~\ref{sec:exploit-explore} into a compact diagnostic guide. We intend it as both a deployment aid and an analytic contribution: the four diagnostics characterize the axes along which agentic search landscapes structurally differ (failure rate, score spread, drift, and refinement variance), and each axis maps to a concrete hyperparameter adjustment. These full-dataset diagnostics are an analytic study of why agentic search landscapes differ, not a deployment requirement: the fixed default configuration needs no pilot and already matches or beats every task-specialized baseline, while the \ours{}$^*$ refinements adjust only one or two hyperparameters for modest gains (typically $0.2$--$1.2$\,pp). When adaptation is desired, the diagnostics are cheap landscape statistics estimable from a small pilot subset, and most are collected simply by running the baseline method (Section~\ref{sec:validation-heavy}), a run practitioners would perform anyway, so they largely reuse existing computation rather than new budget.
}

\begin{table}[t]
\centering
\color{revcolor}
\small
\caption{Compact diagnostic guide: mapping pilot-run signatures to hyperparameter adjustments, condensing Section~\ref{sec:exploit-explore}.}
\label{tab:diagnostic-guide}
\resizebox{\columnwidth}{!}{%
\begin{tabular}{p{2.4cm} p{4.3cm} p{3.3cm} p{2.0cm}}
\toprule
\textbf{Diagnostic signature} & \textbf{What it says about the task} & \textbf{Relevant hyperparameter} & \textbf{Evidence} \\
\midrule
High failure rate $\rho$ (e.g.\ ${>}0.5$) & most expansions fail, so separating productive nodes is critical & success exponent $\alpha$ & HotpotQA \\
\addlinespace
Low $\rho$ (e.g.\ ${<}0.1$) & failures are rare & success exponent $\alpha$ (little effect, keep default) & K-MSE, DROP \\
\addlinespace
High score drift $\kappa$ & early rankings are unreliable as bounds re-rank nodes & exploration type (use UCT) and score gate $\tau$ & HoVeR \\
\addlinespace
Low $\kappa$ & rankings stabilize early, safe to commit & exploration type (use PUCT) & HotpotQA \\
\addlinespace
Low $\hat{\sigma}_{\mathcal{A}}$ and low $\kappa$, or weak validator & aggressive exploitation over-commits to spurious gaps & shaping temperature $T$ & K-MSE, DROP, LCB \\
\bottomrule
\end{tabular}%
}
\end{table}

{\color{revcolor}
The default configuration, used when no pilot is run, is PUCT with $C\!=\!1.0$, $\alpha\!=\!\sqrt{2}$, $T\!=\!0.3$, $M_0\!=\!3$, and $\tau\!=\!0.25$ (Table~\ref{tab:defaults}).
}

\section{{Further Evaluation on an Additional Domain: GPU Kernel Optimization}}
\label{app:gpu-kernel}

{\color{revcolor}
To further probe the generalizability of \ours{}, we evaluate it on an additional domain beyond the four in the main text: GPU kernel optimization. Using the same fixed default configuration (Table~\ref{tab:defaults}), with no tuning and no diagnostics, we follow the setup of \citet{ifco}, using Claude Opus 4.6 as the refinement model and their original configuration for all other settings, and reproduce their pipeline on 40 tasks, running each task on a single NVIDIA A100 40GB GPU, and compare against their optimized MCTS baseline, which we label MCTS (optimized). Owing to the substantial compute cost of this domain, we report results from a single run rather than the multiple runs used elsewhere in the paper. In this run, \ours{} improves over MCTS (optimized) on every metric, raising the geometric-mean speedup from $4.15\times$ to $4.36\times$, the median speedup from $3.09\times$ to $3.81\times$, and the per-category speedups across all three difficulty levels (Table~\ref{tab:gpu-kernel}). We note that the original speed validation of \citet{ifco} contained an issue that admitted invalid (spurious) speedups; the numbers reported here are the corrected results that exclude those invalid speedups. These results indicate that \ours{}'s budget-allocation mechanisms generalize to this additional domain without any domain-specific adaptation.
}

\begin{table}[t]
\caption{{\color{revcolor}Further evaluation on an additional domain, GPU kernel optimization: 40 GPU-kernel tasks, each run on a single NVIDIA A100 40GB GPU, using \ours{}'s fixed default configuration (Table~\ref{tab:defaults}) with no tuning or diagnostics. All values are speedup factors ($\times$); L1, L2, and L3 are the difficulty levels (categories) defined by \citet{ifco}. \ours{} improves over MCTS (optimized) on every metric. Results are from a single run with Claude Opus 4.6, owing to the substantial compute cost of this domain. The reported speedups are the corrected results that exclude the invalid speedups admitted by the original validation of \citet{ifco}.}}
\label{tab:gpu-kernel}
\centering
\color{revcolor}
\small
\resizebox{\columnwidth}{!}{%
\begin{tabular}{lccccc}
\toprule
\textbf{Method} & \textbf{Geo. mean} & \textbf{Median} & \textbf{L1} & \textbf{L2} & \textbf{L3} \\
\midrule
MCTS (optimized) & $4.15\times$ & $3.09\times$ & $3.68\times$ & $5.04\times$ & $3.69\times$ \\
\ours{} & $\mathbf{4.36\times}$ & $\mathbf{3.81\times}$ & $\mathbf{3.76\times}$ & $\mathbf{5.25\times}$ & $\mathbf{4.30\times}$ \\
\bottomrule
\end{tabular}%
}
\end{table}

\section{{Worked Example: The Virtual Child}}
\label{app:virtual-example}

{\color{revcolor}
\paragraph{Setup.} The numbers are taken from a real default-config \ours{} run on HotpotQA (seed~0, Qwen3-8B) and computed exactly as in Equations~\ref{eq:nfair} and~\ref{eq:virtual} and Algorithm~\ref{alg:uct-select}. We reconstruct the iteration where the run's best candidate (score $68.67$) is created: at iteration~104 the search descends $\text{root} \rightarrow v_1 \rightarrow v_3 \rightarrow v_7$ and expands the leaf $v_7$. At that point the global score range is $[50.33, 65.0]$ ($68.67$ does not yet exist, being the child about to be created) and the defaults apply ($T\!=\!0.3$, $\alpha\!=\!\sqrt{2}$, $C\!=\!1.0$, PUCT with parent exponent $0.5$). We show the virtual-child comparison at the pivotal node $v_1$, which has $n_v\!=\!28$ visits, $n_v^+\!=\!4$ successful visits (success rate $0.143$), score $s_v\!=\!52.33$, and a single child $v_3$ ($n\!=\!24$, $n^+\!=\!3$, score $55.33$, rewards $\{62.33, 60.67, 65.0\}$).

\paragraph{Best real child.} The rewards of $v_3$ normalize and shape (Eq.~\ref{eq:shaping}) to $\{0.528, 0.351, 1.000\}$ with mean $0.626$; its success rate $3/24\!=\!0.125$ gives $(0.125)^{\sqrt{2}}\!=\!0.053$, so the exploitation term is $0.053 \cdot 0.626\!=\!0.033$. With PUCT exploration $C\cdot\sqrt{n_v}/(1+n_{\text{child}})\!=\!\sqrt{28}/(1+24)\!=\!0.212$, the total is $\text{ExUCT}(v_3)\!=\!0.245$.

\paragraph{Virtual child.} The fair-share visit count (Eq.~\ref{eq:nfair}) is $n_{\text{fair}}\!=\!24/1\!=\!24$ and the sample size is $k\!=\!\operatorname{round}(n_{\text{fair}}\cdot n_v^+/n_v)\!=\!\operatorname{round}(24 \cdot 0.143)\!=\!3$. The virtual child draws $k\!=\!3$ scores from the parent's pool (reward history plus $s_v$), namely $\{55.33, 62.33, 60.67, 65.0, 52.33\}$ with shaped values $\{0.078, 0.528, 0.351, 1.000, 0.021\}$. Its exploration term is $\sqrt{28}/(1+24)\!=\!0.212$, so in expectation it scores $(0.143)^{\sqrt{2}}\cdot 0.396 + 0.212\!=\!0.237$, just below the best real child at $0.245$.

\paragraph{Decision and interpretation.} The comparison is close, $0.237$ versus $0.245$, and turns on the shaped sample: widening wins only when the mean of the three draws exceeds about $0.52$, which requires repeatedly drawing the two high pool entries ($0.528$ and $1.000$). Enumerating all draws, this happens with probability $0.256$, so \ours{} deepens into $v_3$ about three quarters of the time, and the descent continues through $v_3$ to the leaf $v_7$ whose expansion produces the global best of $68.67$. The virtual child's exploration bonus, computed at the honest fair-share count $n_{\text{fair}}\!=\!24$, does not on average overcome a maturing real child that already carries an exploitation signal, so the mechanism favors deepening the productive chain. A fixed first-play-urgency constant cannot make this comparison, as it sees neither the parent's reward distribution nor the maturity of the existing children. The same close comparison recurs at the root (virtual $0.372$ versus best real $0.381$), while at $v_3$ itself widening is gated off (it is at the branching cap $M_0\!=\!3$ and below the widening visit threshold), so the search simply deepens into $v_3$'s least-visited child $v_7$.
}

\section{Ethical Considerations and Broader Impact}
\label{app:ethics}

\ours{} is a general-purpose tree-search policy that improves budget efficiency for LLM-based agentic search. As a search algorithm, it could in principle accelerate any agent-based system, including potentially harmful ones. However, the method itself does not introduce new capabilities beyond what existing LLM agents already possess; it only improves how evaluation budget is allocated among candidates. All experiments use publicly available benchmarks and models accessed through standard APIs. The molecular elucidation task involves deducing known structures from spectra rather than de novo generation of novel compounds. We do not foresee significant dual-use risks specific to this work beyond those inherent to general LLM agent research.

\section{Artifact Documentation}
\label{app:artifacts}

Table~\ref{tab:artifacts} documents the artifacts used in this work, their licenses, and whether our use is consistent with intended purpose.

\begin{table}[h]
\centering
\small
\caption{Artifacts used, with licenses and intended-use consistency.}
\label{tab:artifacts}
\resizebox{\columnwidth}{!}{%
\begin{tabular}{llll}
\toprule
\textbf{Artifact} & \textbf{Type} & \textbf{License} & \textbf{Consistent Use} \\
\midrule
GEPA & Framework & MIT & Yes \\
TreeQuest & Framework & Apache 2.0 & Yes \\
AFlow & Framework & MIT & Yes \\
K-MSE & Framework & MIT & Yes \\
DSPy & Framework & MIT & Yes \\
vLLM & Inference engine & Apache 2.0 & Yes \\
RDKit & Library & BSD-3-Clause & Yes \\
ColBERTv2 & Model & MIT & Yes \\
Qwen3-8B & Model & Apache 2.0 & Yes \\
HotpotQA & Dataset & CC BY-SA 4.0 & Yes \\
HoVeR & Dataset & MIT & Yes \\
DROP & Dataset & CC BY-SA 4.0 & Yes \\
LiveCodeBench & Benchmark & MIT & Yes \\
MolPuzzle & Dataset & MIT & Yes \\
\bottomrule
\end{tabular}%
}
\end{table}

All frameworks are used as search harnesses into which we integrate \ours{}, consistent with their intended purpose as research tools. Datasets and benchmarks are used for evaluation as intended by their creators. Models are accessed via their supported APIs or inference engines.

\section{Use of AI Assistants}
\label{app:ai-use}

We used AI assistants to refine the grammar of the paper's text.

\end{document}